\documentclass{article}

\usepackage[square,numbers]{natbib}

\usepackage[final]{neurips_2022}

\usepackage[utf8]{inputenc} 
\usepackage[T1]{fontenc}    
\usepackage[hypertexnames=false]{hyperref}       
\usepackage{url}            
\usepackage{booktabs}       
\usepackage{amsfonts}       
\usepackage{bm}
\usepackage{nicefrac}       
\usepackage{microtype}      
\usepackage{xcolor,cancel}         
\usepackage{amsmath}

\usepackage[all]{hypcap}
\usepackage{graphicx}
\graphicspath{ {./figures/} }
\usepackage{subfiles}
\usepackage{xr} 
\usepackage{calc}
\newsavebox\CBox
\newcommand\hcancel[2][1pt]{%
  \ifmmode\sbox\CBox{$#2$}\else\sbox\CBox{#2}\fi%
  \makebox[0pt][l]{\usebox\CBox}%
  \rule[0.2\ht\CBox-#1/2]{\wd\CBox}{#1}}
  
\newcommand{\beginsupplement}{%
        \setcounter{table}{0}
        \renewcommand{\thetable}{S\arabic{table}}%
        \setcounter{figure}{0}
        \renewcommand{\thefigure}{S\arabic{figure}}%
     }

\title{Distinguishing Learning Rules with Brain Machine Interfaces}

\author{
  Jacob P. Portes \\
  Center for Theoretical Neuroscience\\
  Columbia University\\
  \texttt{j.portes@columbia.edu} \\
  \And
  Christian Schmid \\
   Institute of Neuroscience \\
   University of Oregon \\
   \texttt{cschmid9@uoregon.edu} \\
  \AND
  James M. Murray\thanks{corresponding author} \\
    Institute of Neuroscience \\
   University of Oregon \\
   \texttt{jmurray9@uoregon.edu} \\
}

\begin{document}
\maketitle


\begin{abstract}
Despite extensive theoretical work on biologically plausible learning rules, clear evidence about whether and how such rules are implemented in the brain has been difficult to obtain. We consider biologically plausible supervised- and reinforcement-learning rules and ask whether changes in network activity during learning can be used to determine which learning rule is being used. Supervised learning requires a credit-assignment model estimating the mapping from neural activity to behavior, and, in a biological organism, this model will inevitably be an imperfect approximation of the ideal mapping, leading to a bias in the direction of the weight updates relative to the true gradient. Reinforcement learning, on the other hand, requires no credit-assignment model and tends to make weight updates following the true gradient direction. We derive a metric to distinguish between learning rules by observing changes in the network activity during learning, given that the mapping from brain to behavior is known by the experimenter. Because brain-machine interface (BMI) experiments allow for precise knowledge of this mapping, we model a cursor-control BMI task using recurrent neural networks, showing that  learning rules can be distinguished in simulated experiments using only observations that a  neuroscience experimenter would plausibly have access to. 
\end{abstract}




\section{Introduction}

In order to update synaptic weights effectively during learning, a biological or artificial neural network must solve the problem of credit assignment. That is, a neuron must infer whether it should increase or decrease its activity in order to decrease the error in the behavior that it is driving.
Supervised learning (SL) provides one approach for solving the credit assignment problem by endowing the synaptic learning rule with a \textit{credit assignment mapping}, i.e.~an estimate of how each neuron's activity affects the behavioral readout. Indeed, gradient-based methods such as backpropagation for feedforward networks and backpropagation through time for recurrent networks do exactly this by multiplying a (possibly multi-dimensional) readout error by the downstream readout weights in order to update hidden-layer synaptic weights.
However, because neurons in a biological neural network likely lack complete information about their downstream projections, ideal credit assignment mappings such as those used in gradient-based methods are biologically implausible--an issue known as the weight transport problem \cite{grossberg1987competitive,richards2019deep,lillicrap2020backpropagation}.
Recent work has shown that effective supervised learning in both feedforward networks and RNNs can still take place as long as there is a positive overlap between the true readout weights and the credit assignment matrix used for learning, providing one possible solution to this problem \cite{lillicrap2016random,lillicrap2020backpropagation,murray2019local,akrout2019deep,amit2019deep}.

Reinforcement learning (RL) is another approach for solving the credit assignment problem. Node perturbation algorithms correlate noise injections with scalar reward signals in order to update network weights \cite{williams1992simple,werfel2003learning,miconi2017biologically}. Such algorithms avoid the weight transport problem, as the update rule does not explicitly require a credit assignment matrix mapping the vector error back to the recurrent weights. While the weight updates for algorithms in this family tend to be noisy, the policy-gradient theorem guarantees that they tend to follow the true gradient of the objective function (Appendix \ref{appendix:RL}, \cite{williams1992simple,sutton2018reinforcement}).

Brain-machine interfaces (BMI) provide an ideal paradigm for investigating questions about learning and credit assignment in real brains, as they enable the experimenter to define the mapping (the ``decoder'') from neural activity onto behavior. For example, monkeys can be trained to control a cursor using a BMI in the motor cortex, and the behavior can be relearned following different manipulations of the BMI decoder in the course of a single day \cite{sadtler2014neural,zhou2019distinct,losey2022learning} or over the course of multiple days \cite{athalye2017emergence,oby2019new,sorrell2021brain}. 
In the case of BMIs, the weight transport problem is particularly acute, since the readout weights mapping the neural activity onto the cursor position may be changed abruptly by the experimenter, and there is no plausible way for the agent to generate an instantaneous estimate of the new readout weights in order to then apply a learning rule.
If an agent is to use a credit-assignment model at all in such a task, it must consist of a weight matrix that imperfectly approximates the readout mapping, i.e.~a \textit{biased} internal model of credit assignment.
We refer to this as the \textit{decoder alignment problem}.

The key insight in our work is that, under an SL rule, the agent will invariably perform credit assignment using a biased credit-assignment model, whereas no such model is required under a policy gradient-like RL rule.
Using this insight, we develop a framework for modeling BMI experiments with recurrent neural networks (RNNs) using biologically plausible versions of SL and RL. 
We show that an imperfect credit-assignment model introduces a systematic bias in the weight updates, as well as the direction in which neural activity evolves during learning. Under the assumption that SL is biased but RL is not, this therefore provides a means by which to distinguish these different learning strategies.
We derive a statistical metric for comparing observed changes in neural activity with the changes predicted assuming knowledge of the BMI decoder, and we show how this metric can be used to detect learning bias in our simulated experiments.
While we focus on distinguishing biased SL from unbiased RL, the approach described here could be used more generally to distinguish different biased vs.~unbiased learning rules---or even biased learning rules with two different types of bias---from one another.

\textbf{Related work.} While there has recently been a renaissance of research on biologically plausible learning rules \cite{lillicrap2016random,miconi2017biologically,akrout2019deep,lansdell2019learning,murray2019local,lillicrap2020backpropagation}, there has been relatively little work on how such proposals might be experimentally verified in neural experiments. 
\citet{lim2015inferring} inferred hyperparameters of a Hebbian learning rule based on neural activity from a visual cortical area in monkeys. Their work, however, assumed a single class of learning rules rather than trying to distinguish between multiple classes.
\citet{nayebi2020identifying} showed that different optimizers and hyperparameters used to train convolutional neural networks can in principle be distinguished by a classifier based on both weight changes and neural activations. However, the classifier was trained using ground-truth data about the identity of the learning rule, which is the very thing that we aim to infer in our model. 
\citet{kepple2021curriculum} proposed a curriculum-based method for discerning the objective being optimized with SL and showed via RNN modeling that different objectives lead to distinct task-acquisition timescales. While their approach analyzed evidence accumulation and decision tasks and focused on distinguishing different objectives, we focus primarily on continuous control tasks and on distinguishing learning rules. While 
\citet{feulner2021neural} simulated BMI learning with RNNs to compare learning that required changes within vs.~outside of the intrinsic manifold of neural activity, and \citet{feulner2022small} used a similar approach to analyze how learning at upstream or recurrent synapses affected the covariance of neural activity, neither of these studies attempted to distinguish learning rules.
To our knowledge, our study is the first to propose a framework for distinguishing between learning rules using BMIs.


\begin{figure}[t!]
  \centering
  \includegraphics[width=1\textwidth]{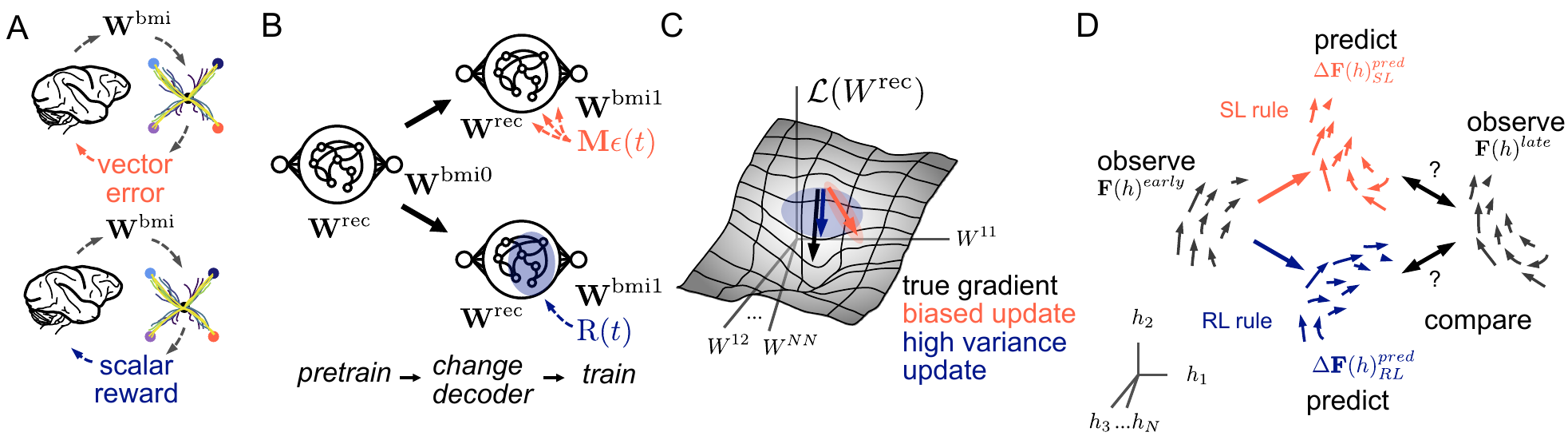}
  \caption{Approach. (A) We model a brain-machine interface experiment in which a monkey learns to move a cursor on a screen using either supervised learning (SL, top) or reinforcement learning (RL, bottom).
  (B) We pretrain an RNN to perform a center-out reach task with a fixed decoder $\mathbf{W}^{\mathrm{bmi0}}$, switch the decoder to $\mathbf{W}^{\mathrm{bmi1}}$, then train the recurrent weights $\mathbf{W}^{\mathrm{rec}}$ again with either SL or RL. 
  (C) SL with a biased credit assignment mapping will lead to biased recurrent weight updates, whereas RL leads to noisy but unbiased weight updates. 
  (D) By inferring the neural activity flow fields from observed data and comparing the observed change during learning with the predicted changes assuming SL or RL, we infer which learning rule was used to train the RNN.}
 \label{fig:intro}
\end{figure}

\section{Theoretical approach}
\label{sec:theory}
The overall approach for our work is shown schematically in Fig.~\ref{fig:intro}.
We train RNNs to perform a BMI-inspired cursor-control task using different learning rules, with the goal of analyzing the observed activity to infer the learning rule that was used to train the network.
Inspired by numerous BMI experiments \cite{sadtler2014neural,golub2018learning,oby2019new}, we pre-train the RNN using a particular choice of decoder weights, then require the RNN to relearn the task using a new set of decoder weights. 
Our main result is to predict the different ways in which the RNN activity is expected to evolve through learning, then to compare these predictions with the observed changes in the activity in order to infer the learning rule.
In the sections that follow, we first introduce the SL and RL update rules that are used for training our networks. We then show that they lead to distinct predictions about how the RNN activity is expected to evolve with training. The remainder of the paper presents empirical results showing that the learning rules can be inferred by observing the network activity, first for a relatively simple case and then under more generalized assumptions.

While our main results can be obtained using either a feedforward or recurrent circuit model, we mostly focus on the latter case because BMI experiments typically involve control tasks driven by time-varying activity in motor cortex, for which an RNN model is more appropriate. The feedforward version of our results is provided in Appendix \ref{appendix:feedforward}.


\subsection{Modeling BMI learning with RNNs} 
We model motor cortex as a vanilla RNN characterized by input weights $\mathbf{W}^{\mathrm{in}}$, recurrent weights $\mathbf{W}^{\mathrm{rec}}$, feedback weights $\mathbf{W}^{\mathrm{fb}}$, and BMI decoder weights $\mathbf{W}^{\mathrm{bmi}}$. Since the experimenter has complete control over the decoder in a BMI experiment, we treat $\mathbf{W}^{\mathrm{bmi}}$ as fixed and assume that learning occurs in recurrent weights $\mathbf{W}^{\mathrm{rec}}$ alone. 
The activity of the recurrent units $\mathbf{h}^t$ is given by
\begin{equation}\label{eq:update}
    h_i^t = \Big(1- \frac{1}{\tau} \Big) h_i^{t-1} + \frac{1}{\tau}  \phi \Big( \sum_j W^{\mathrm{rec}}_{ij} h_j^{t-1} + \sum_j W^{\mathrm{in}}_{ij} x_j^t  + \sum_j W^{\mathrm{fb}}_{ij} y_j^{t-1}  \Big) + \xi_i^t,
\end{equation}
where $\tau$ is the RNN time constant, $\phi(\cdot) =\tanh(\cdot)$ is a nonlinear activation function, $\mathbf{x}^t$ is an input signal, and $\xi_i^t \sim \mathcal{N}(0,\sigma^2_{\mathrm{rec}})$ is isotropic noise added to the recurrent units. This activity is read out by $\mathbf{W}^\mathrm{bmi}$ to obtain the cursor position $ y_k^t = \sum_i W^{\mathrm{bmi}}_{ki} h_i^t$. The error at each time step is defined as $\epsilon^t_k = {y^*}_k^t - y_k^t$, where $\mathbf{y^*}^t$ is the target output for each time step. The loss to be minimized is then $L = \frac{1}{2T} \sum_{t=1}^T \sum_{k=1}^{N_y} ( \epsilon_k^t)^2$ where $T$ is the trial duration and $N_y=2$ is the number of readout dimensions.

\textbf{Supervised learning.} 
Standard supervised algorithms for RNNs such as Backpropagation Through Time \cite{rumelhart1985learning} and Real-Time Recurrent Learning \cite{williams1989learning} are widely acknowledged to be biologically implausible. 
The reasons for this are two-fold. First, both algorithms assume knowledge of the readout weights for purposes of credit assignment, which we referred to above as the weight transport problem. Second, updating a particular recurrent weight with one of these algorithms requires knowledge of all of the other weights in the network, which is not information that a biological neuron would likely have access to.

A recently proposed alternative, Random Feedback Local Online (RFLO) learning \cite{murray2019local}, is an approximate gradient-based algorithm with $\Delta \mathbf{W}^\mathrm{rec} \approx -\partial L/\partial \mathbf{W}^\mathrm{rec}$ that addresses these problems by (i) assuming that error information is projected back into the network with weights that imperfectly approximate the ideal credit assignment mapping, and (ii) dropping nonlocal terms from the update rule.
The recurrent weight update with this learning rule is
\begin{equation}\label{eq:SL_update}
    \Delta W_{ij}^\mathrm{rec} = 
    \eta \sum_{t=1}^T 
    \left[\mathbf{M} \boldsymbol{\epsilon}^t\right]_i
    p_{ij}^t, \;\; \;\;  \;\;  \;\;  \;\;  \;\;  \;\;   p_{ij}^t = 
    \left( 1 - \frac{1}{\tau} \right) p_{ij}^{t-1} 
    + \frac{1}{\tau} \phi'(u_i^t) h_j^{t-1},
\end{equation}
with $u_i^t = \sum_k W^{\mathrm{rec}}_{ik} h_k^{t-1} + \sum_k W^{\mathrm{in}}_{ik} x_k^t  + \sum_k W^{\mathrm{fb}}_{ik} y_k^t $, and $p_{ij}^t \approx \partial h_i^t /\partial W_{ij}^\mathrm{rec}$.
Further details are provided in Appendix \ref{appendix:SL}. The learning rule that we implement differs slightly from RFLO in that the readout weights, as in a BMI experiment, are fixed during training rather than learned. In RFLO, learning of the readout weights was required in order to align the readout weights with the credit assignment matrix. Here we will instead assume that the credit assignment matrix $\mathbf{M}$ is fixed and partially aligned with $(\mathbf{W}^\mathrm{bmi})^\top$, so that the cosine similarity of the flattened matrices satisfies $\mathrm{sim}(\mathbf{M}, (\mathbf{W}^\mathrm{bmi})^\top) > 0$. 

In \eqref{eq:SL_update}, the credit assignment mapping $M_{ik}$ assigns responsibility for the readout error $\epsilon^t_k$ to neuron $i$. 
While the ideal mapping would be $\mathbf{M} = (\mathbf{W}^\mathrm{bmi})^\top$, learning can still be effective as long as there is a positive alignment between $\mathbf{M}$ and $(\mathbf{W}^\mathrm{bmi})^\top$ \cite{murray2019local} (as shown previously in the case of feedforward networks \cite{lillicrap2016random}).
In the case where they are not equal, as we presume they cannot be in a BMI experiment, the matrix $\mathbf{M}$ constitutes a \textit{biased} credit-assignment mapping. Below, we show that this leads to a predictable bias in the change of the RNN activity as the task is learned.

\begin{figure}[t!]
  \centering
  \includegraphics[width=1\textwidth]{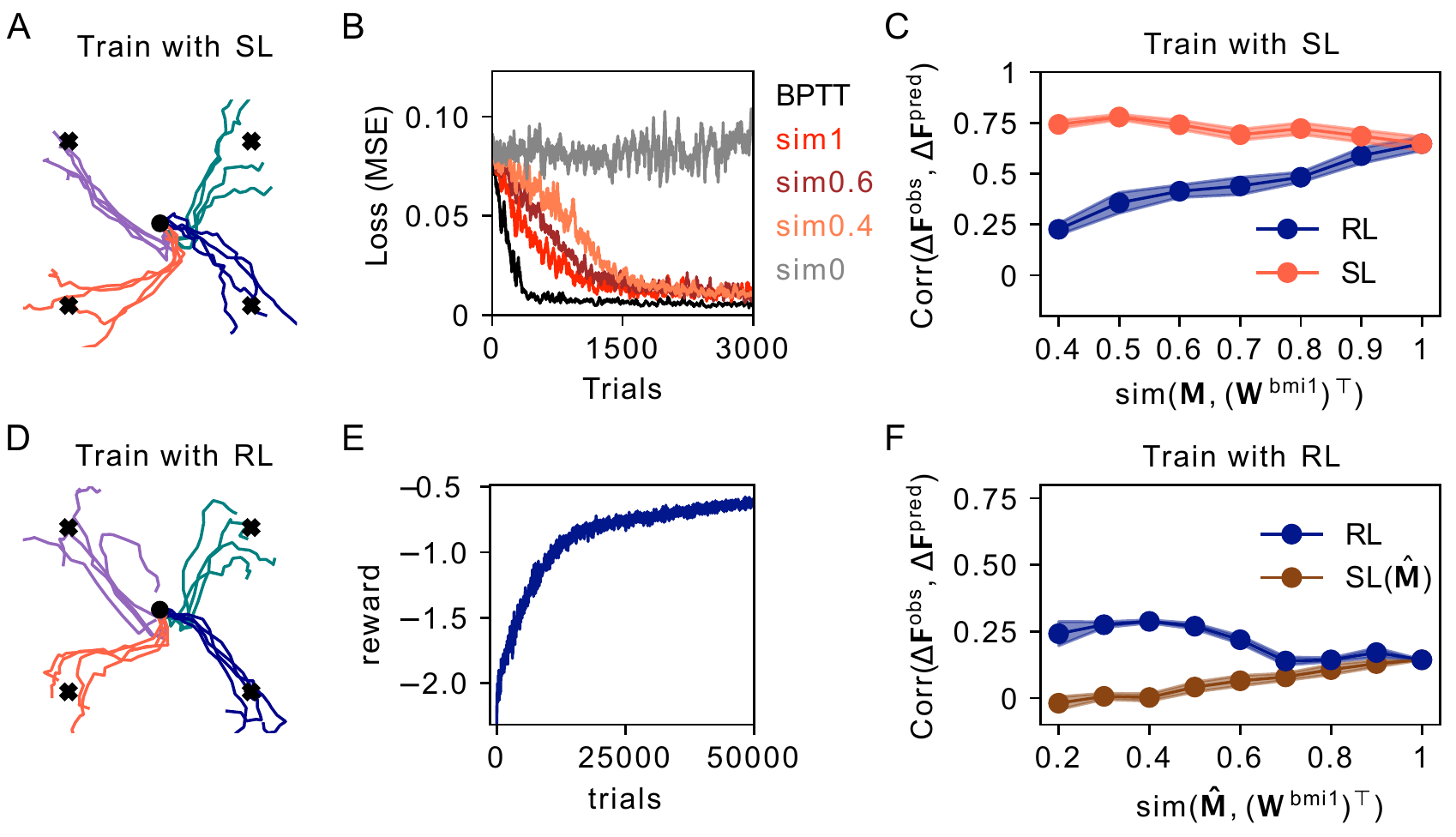}
  \caption{Distinguishing learning rules in a trained network. 
  (A) Example output traces from an RNN trained with SL to move a cursor to one of four targets.
  (B) When trained with SL, the similarity of the credit assignment mapping with the readout weights determines the speed of training (red curves). Learning in an RNN trained with backpropagation through time is shown for comparison (black curve).
  (C) For an RNN trained with SL, the similarity of the observed flow field change with the predicted change assuming either SL (red) or RL (blue).
  (D) Example output traces from an RNN trained with RL to move a cursor to one of four targets.
  (E) Learning curve for an RNN trained with RL.
  (F) For an RNN trained with RL, the similarity of the observed flow field change with the predicted change assuming either SL with a randomly sampled $\mathbf{\hat{M}}$ (brown) or RL (blue).}
 \label{fig:main-results}
\end{figure}

Example output trajectories from an RNN trained on the cursor-control task with the SL rule described above are shown in Fig.~\ref{fig:main-results}A. The results in Fig.~\ref{fig:main-results}B show that learning the task is possible whenever there is partial alignment between $\mathbf{M}$ and  $(\mathbf{W}^\mathrm{bmi})^\top$, and that faster learning occurs as this alignment increases.

\textbf{Reinforcement learning.} To train RNNs with RL, we use a node perturbation learning rule closely related to REINFORCE \cite{williams1992simple}. We first define a scalar reward $R^t=-|\boldsymbol{\epsilon}^t|^2$. 
Node perturbation seeks to increase this reward by trial and error using the noise appearing in the recurrent population activity in \eqref{eq:update}. If the noise to a given neuron causes the reward to increase, then the weights will be updated so that that neuron's input will be greater in subsequent trials. Including an eligibility trace in the learning rule further makes it possible to assign credit for changes in reward to noise events at previous time steps. Finally, we can subtract off a baseline from the reward so that learning is guided by a reward prediction error, which somewhat decreases the variance of the weight updates without introducing bias \cite{sutton2018reinforcement}. The resulting learning rule, which we derive using the policy gradient theorem \cite{marbach2001simulation,sutton2018reinforcement} in Appendix \ref{appendix:RL}, is
\begin{equation}
    \label{eq:RL_update}
    \Delta W_{ij}^{\mathrm{rec}} = \eta \sum_t^T (R^t - \bar{R}^t) q_{ij}^t,  \;\;  \;\;  \;\;  \;\;  \;\;  \;\;  \;\;  \;\;  q_{ij}^t =
    \left( 1 - \frac{1}{\tau} \right) q_{ij}^{t-1} 
    + \frac{1}{\tau} \xi_i^t \phi'(u_i^t) h_j^{t-1}.
\end{equation}
Very similar RL rules for RNNs have been previously proposed and studied in Refs.~\cite{fiete2006gradient,miconi2017biologically}. Importantly for our purposes, the fact that this learning rule is derived using a policy gradient approach guarantees that, upon averaging over isotropic noise, the weight updates follow the true gradient of the loss function (in our case given by $L = -\frac{1}{2} \sum_t R^t$), making this an \textit{unbiased} learning rule. Like \eqref{eq:SL_update}, \eqref{eq:RL_update} does not have nonlocal terms. Example output trajectories and the learning curve from an RNN trained on the cursor-control task using \eqref{eq:RL_update} are shown in Fig.~\ref{fig:main-results}D,E.


\subsection{Characterizing changes in neural activity with vector flow fields} 
Because observing changes in neural activity is much more straightforward than observing changes in synaptic weights in neuroscience experiments, we aim to identify experimentally observable signatures in the neural activity that might distinguish biased SL from unbiased RL.
Because the RNN activity consists of time-dependent trajectories, and changes in the weights will generally affect entire trajectories cumulatively from one time step to the next, calculating the change $\Delta \mathbf{h}^t$ due to an update $\Delta \mathbf{W}^\mathrm{rec}$ is not entirely straightforward (though \textit{cf.}~the feedforward case described in Appendix \ref{appendix:feedforward}, where this can be done). 
A more useful approach is to quantify the learning-induced change in the vector flow field that shapes the RNN activity, where the flow field can be estimated empirically by observing neural activity trajectories.
To do this, we linearize \eqref{eq:update}, assuming $\phi(u) = u$, to obtain an expression for a vector flow field function $\mathbf{F}(\mathbf{h})$ that maps activity $\mathbf{h}^t \longrightarrow \mathbf{h}^{t+1}$ and is defined for all points $\mathbf{h}$ in neural activity space:
\begin{equation}\label{eq:flowfield}
\mathbf{F}({\mathbf{h}}) 
= \frac{1}{\tau} \left[ (\mathbf{W}^{\mathrm{rec}} - \mathbb{I}) \mathbf{h} + \mathbf{W}^\mathrm{in} \mathbf{x} + \mathbf{W}^\mathrm{fb} \mathbf{y} \right],
\end{equation}
where, in our simulations, $\Delta t = 1$. Thus, from \eqref{eq:update}, we see that $\mathbf{F}(\mathbf{h}^t) \approx \mathbf{h}^{t+\Delta t} - \mathbf{h}^t$.
If the recurrent weights change according to $\mathbf{W}^{\mathrm{rec}} \rightarrow \mathbf{W}^{\mathrm{rec}} + \Delta \mathbf{W}^{\mathrm{rec}}$, then the flow field exhibits a corresponding change $\mathbf{F} \rightarrow \mathbf{F} + \Delta \mathbf{F}$, where $\Delta \mathbf{F}({\mathbf{h}}) = \frac{1}{\tau} \Delta \mathbf{W}^{\mathrm{rec}} \mathbf{h}.$
This equation enables us to relate changes in $\mathbf{W}^\mathrm{rec}$ to changes in neural activity (note that this assumes that plasticity is occurring in the recurrent weights $\mathbf{W}^{\mathrm{rec}}$ alone, and that $\mathbf{W}^{\mathrm{in}}$ and $\mathbf{W}^{\mathrm{fb}}$ do not change). 
As illustrated in Fig.~\ref{fig:intro}D, our approach will be to (i) predict the expected change in flow field $\Delta\mathbf{F}^{\mathrm{pred}}(\mathbf{h})$ under either SL or RL, (ii) estimate $\Delta\mathbf{F}^{\mathrm{obs}}(\mathbf{h})$ empirically using activity observed early and late in learning, and (iii) determine which learning rule was used to train the network by comparing the empirically observed vs.~predicted flow field changes.


In order to predict the flow field change in the SL case for a point $\mathbf{h}$ in neural activity space, we predict the direction of weight change using an approximation of \eqref{eq:SL_update}:
\begin{equation}
    \label{eq:W_pred_SL}
    \Delta \mathbf{F}^{\mathrm{pred}}_{\mathrm{SL}}(\mathbf{h}) = \Delta \mathbf{W}^{\mathrm{pred}} |_{\mathrm{SL}} \mathbf{h},  \;\;  \;\;  \;\;  \;\;  \;\;  \;\;  \;\;  \;\; \Delta W_{ij}^{\mathrm{pred}}|_{\mathrm{SL}} =  \sum_{n \in \mathrm{mid}} \sum_t \sum_k M_{ik} \epsilon_k^{n,t}  h^{n,t}_j,
\end{equation}
where $n$ refers to trials, and $n\in\mathrm{mid}$ denotes that we sample from trials during the middle part of training, when most of the learning is presumed to take place.
In the RL case, we similarly predict the direction of weight change using an approximation of \eqref{eq:RL_update}:
\begin{equation}
    \label{eq:W_pred_RL}
    \Delta \mathbf{F}^{\mathrm{pred}}_{\mathrm{RL}}(\mathbf{h}) = \Delta \mathbf{W}^{\mathrm{pred}} |_{\mathrm{RL}} \mathbf{h},  \;\;  \;\;  \;\;  \;\;  \;\;  \;\;  \;\;  \;\; \Delta W_{ij}^{\mathrm{pred}}|_{\mathrm{RL}} =  \sum_{n\in\mathrm{mid}} \sum_t \sum_k W^{\mathrm{bmi}}_{kl} \Sigma_{il} \epsilon_k^{n,t}  h^{n,t}_j.
\end{equation}
In this equation, $\boldsymbol{\Sigma}$ is the recurrent noise covariance, which, under conditions of isotropic noise, reduces to a prefactor $\sigma^2_{\mathrm{rec}} \mathbb{I}$. 

Importantly for our purposes, the predicted changes in flow field from \eqref{eq:W_pred_SL} and \eqref{eq:W_pred_RL} depend entirely on quantities that are known or can be estimated from experimental data, including $\mathbf{W}^\mathrm{bmi}$, as well as samples of $\mathbf{h}^{n,t}$ and $\boldsymbol{\epsilon}^{n,t}$. As shown in Section \ref{sec:estimating-credit-assignment} below, $\mathbf{M}$ can also be estimated from data. Because we are asking only about changes in the flow field rather than the flow field itself, these quantities do not depend explicitly on the recurrent connectivity $\mathbf{W}^\mathrm{rec}$ itself, which is presumably unobtainable experimentally.
The key distinction between \eqref{eq:W_pred_SL} and \eqref{eq:W_pred_RL} is that the change in flow field in the SL case depends on $\mathbf{M}$, while the change in flow field in the RL case depends on $(\mathbf{W}^{\mathrm{bmi}})^\top$. 
We motivate \eqref{eq:W_pred_SL} and \eqref{eq:W_pred_RL} by computing $\langle \Delta \mathbf{W}^\mathrm{rec} \rangle$ for each learning rule (averaging over noise) in Appendix \ref{appendix:SL} and Appendix \ref{appendix:RL}, respectively. 
We provide empirical evidence that updates approximately follow a consistent direction for each rule in Appendix \ref{sec:update_direction}. 




Next, we wish to estimate the empirical change in the flow field given neural activity observed at different points in learning. 
The approach is somewhat simplified if we assume that the trials during which learning takes place are preceded and followed by ``early'' and ``late'' blocks of trials, in which the performance is observed but the weights are not updated. The assumption that the amount of learning during these early and late blocks is negligible is justified in the limit where the learning rate is small and the number of training trials is large. (Evidence that our main results do not depend strongly on this assumption is given in Appendix \ref{appendix:feedforward}, where learning was allowed to take place during the early and late blocks.)
Then, from \eqref{eq:flowfield}, we define the observed flow field from time step $t$ in trial $n$ as either $\mathbf{F}^\mathrm{early}(\mathbf{h}) = (\mathbf{A}^\mathrm{early} - \mathbb{I})\mathbf{h}$ or $\mathbf{F}^\mathrm{late}(\mathbf{h}) = (\mathbf{A}^\mathrm{late} - \mathbb{I})\mathbf{h}$, where $\mathbf{A}^\mathrm{early}$ is the least-squares solution to the autoregression equation $\mathbf{h}^{n,t+1} = \mathbf{A}^\mathrm{early} \mathbf{h}^{n,t}$ for all time steps $t$ in all early trials $n$, and $\mathbf{A}^\mathrm{late}$ is the corresponding solution for late trials.
The empirical change in the flow field due to learning for any point $\mathbf{h}$ is then given by
$\Delta \mathbf{F}^\mathrm{obs} (\mathbf{h})
= \mathbf{F}^\mathrm{late} (\mathbf{h}) - 
\mathbf{F}^\mathrm{early} (\mathbf{h})$.

Having defined the observed and predicted changes in the flow fields for arbitrary $\mathbf{h}$, we can define the correlation between these quantities as 
\begin{equation}\label{eq:dF_corr}
    \mathrm{Corr}(\Delta \mathbf{F}^{\mathrm{obs}},\Delta \mathbf{F}^{\mathrm{pred}}) = \frac{1}{N_\mathrm{trials} T} 
    \sum_{n=1}^{N_\mathrm{trials}} \sum_{t=1}^T 
    \frac{\Delta \mathbf{F}^{\mathrm{obs}} (\mathbf{h}^{n,t}) \cdot 
    \Delta \mathbf{F}^{\mathrm{pred}}(\mathbf{h}^{n,t})}
    {|\Delta \mathbf{F}^{\mathrm{obs}} (\mathbf{h}^{n,t})| \;  
    |\Delta \mathbf{F}^{\mathrm{pred}} (\mathbf{h}^{n,t})|}.
\end{equation}
We refer to this as the flow field change correlation (FFCC).
One important point to note is that, because this quantity is computed only at observed values of $\mathbf{h}^{n,t}$, it is not the case that the flow field needs to be estimated everywhere in the high-dimensional space of neural activity. If the activity tends to occupy a low-dimensional manifold within this space, then the flow field needs to be determined only within this manifold. Another point to note is that this metric is independent of the magnitudes of $\Delta \mathbf{F}^\mathrm{obs}$ and $\Delta \mathbf{F}^\mathrm{pred}$, so that it does not depend on overall prefactors such as the learning rates and network time constant. In the simulations that follow, the correlation metric \eqref{eq:dF_corr} will be computed separately under our two hypothesized learning rules, i.e.~with $\Delta \mathbf{F}^{\mathrm{pred}} = \Delta \mathbf{F}^{\mathrm{pred}}_\mathrm{SL}$ and with $\Delta \mathbf{F}^{\mathrm{pred}} = \Delta \mathbf{F}^{\mathrm{pred}}_\mathrm{RL}$. 
While the absolute magnitudes of this quantity are not informative on their own, we show below that the relative values can be compared to infer which of the two learning rules is more likely to have generated the data.

\section{Simulation results}\label{sec: results}
In this section, we use the theory outlined above to show that the SL and RL rules we consider can be distinguished in simulations, first in a basic case with simplifying assumptions, and then in more realistic cases where these assumptions are relaxed.

\subsection{Distinct neural signatures of biased SL and unbiased RL} In order to emulate monkey BMI experiments (e.g., \cite{sadtler2014neural,oby2019new,athalye2017emergence,zhou2019distinct,losey2022learning}), we first pretrain the recurrent weights $\mathbf{W}^{\mathrm{rec}}$ of an RNN with  fixed, random readout weights $\mathbf{W}^{\mathrm{bmi0}}$ to perform a center-out cursor-control task, in which a cursor must be moved to one of four target locations specified by the input to the RNN. 
We then change the BMI decoder to $\mathbf{W}^{\mathrm{bmi1}}$, make identical copies of the network, and train one with SL using credit-assignment mapping $\mathbf{M}$ and the other with RL. For these simulations, shown in Fig.~\ref{fig:main-results}, we make the following simplifying assumptions (to be generalized in later sections): (i) the credit assignment mapping $\mathbf{M}$ is fixed and known; (ii) the feedback weights $\mathbf{W}^\mathrm{fb}$ are zero; and (iii) the noise $\boldsymbol{\xi}^t \sim \mathcal{N}(0, \sigma^2_{\mathrm{rec}} \mathbb{I})$ is isotropic.

For the networks trained with biased SL and $\mathrm{sim}(\mathbf{W}^{\mathrm{bmi}0}, \mathbf{W}^{\mathrm{bmi}1}) = 0.5$, we find that a comparison of the observed and predicted flow field changes correctly detects the bias (Fig.~\ref{fig:main-results}C). 
As we vary the alignment between $\mathbf{M}$ and $(\mathbf{W}^{\mathrm{bmi}1})^\top$, we find that the FFCC correctly detects the bias present during learning whenever these two matrices are sufficiently distinct. For the networks trained with RL, we find that the change in activity is aligned more closely to the prediction using true decoder $\mathbf{W}^{\mathrm{bmi}1}$ than when using a random matrix $\mathbf{\hat{M}}$ that is partially aligned to $\mathbf{W}^{\mathrm{bmi}1}$. As we increase the alignment $\mathrm{sim}(\mathbf{\hat{M}}, (\mathbf{W}^{\mathrm{bmi}1})^\top)$, this distinction is less discernible (Fig.~\ref{fig:main-results}F). 
We show in Appendix \ref{sec:exp_details} that these results are robust to changes in hyperparameters.

In the empirical results that follow, we will generalize the assumptions made thus far and show that there is a significant range of parameters where the results still hold. In particular, we will consider the roles played by (i) estimation and/or learning of the credit-assignment model $\mathbf{M}$, (ii) driving feedback via the weights $\mathbf{W}^{\mathrm{fb}}$, and (iii) non-isotropic noise.
\begin{figure}[t!]
  \centering
  \includegraphics[width=1.0\textwidth]{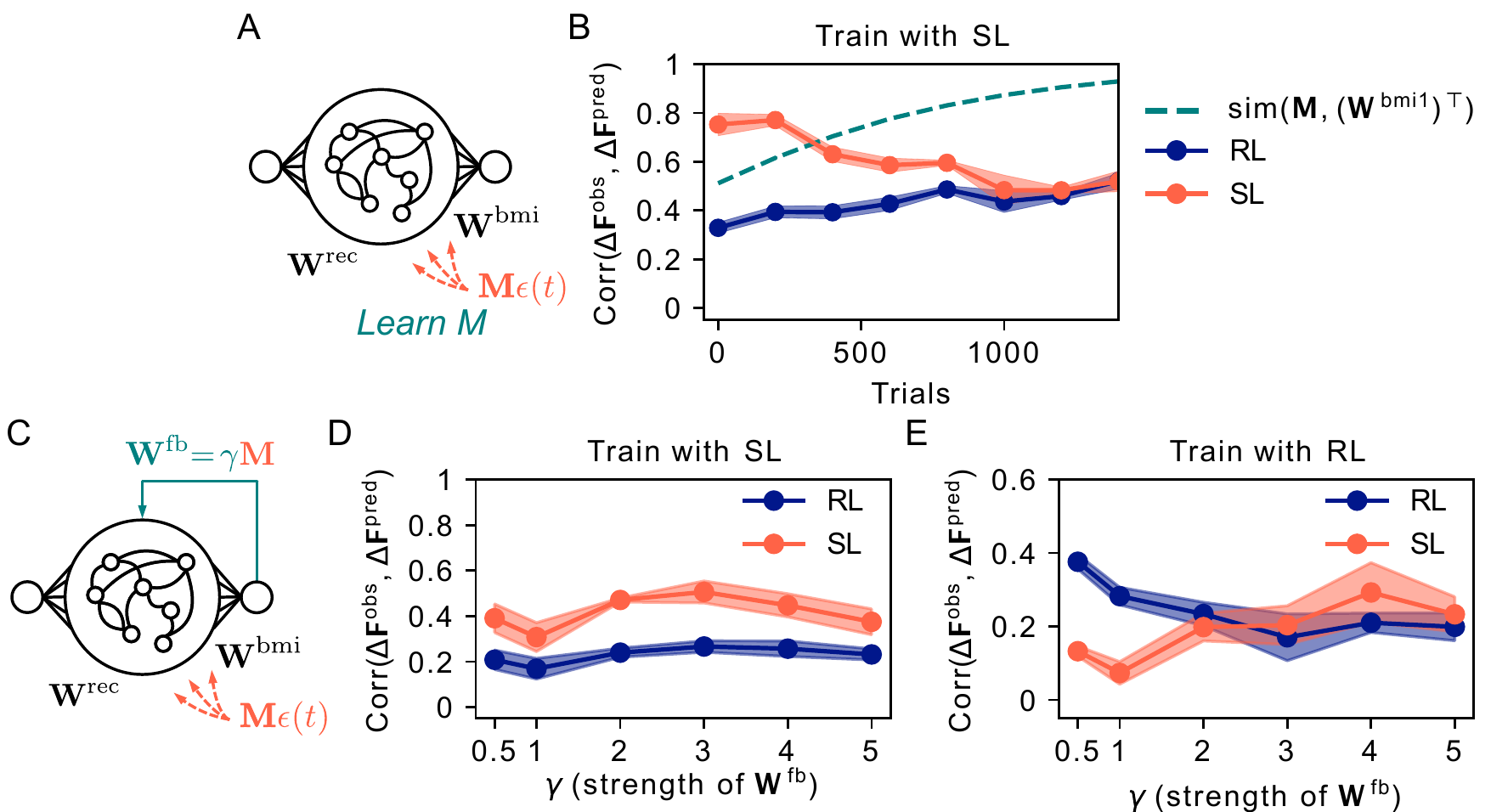}
  \caption{Learning the credit assignment mapping and driving with visual feedback.
  (A) In an RNN trained with SL, the credit assignment mapping $\mathbf{M}$ is learned online during training.
  (B) The similarity of $\mathbf{M}$ to $(\mathbf{W}^\mathrm{bmi})^\top$ increases with training due to learning of $\mathbf{M}$ (green). The flow field metric correctly identifies SL as the learning rule early in training when $\mathbf{M}$ and $\mathbf{W}^\mathrm{bmi}$ are dissimilar, but not later in training when they are very similar.
  (C) An RNN (now with fixed $\mathbf{M}$) is additionally driven by a feedback signal via weights $\mathbf{W}^\mathrm{fb} \propto \mathbf{M}$.
  (D) Comparison between predicted and observed changes in the activity flow field as a function of feedback gain for an RNN trained with SL.
  (E) Comparison between predicted and observed changes in the activity flow field as a function of feedback gain for an RNN trained with RL.}
    \label{fig:feedback}
\end{figure}

\subsection{Learning the credit assignment mapping} In order to optimize learning in a supervised setting, a learner's credit assignment mapping could improve over time to align more closely with the decoder (Fig.~\ref{fig:feedback}A). To investigate the effect of such learning in our model, we applied our analysis to an RNN in which recurrent weights $\mathbf{W}^{\mathrm{rec}}$ were trained with SL, with a biased credit assignment mapping $\mathbf{M}$ that slowly improved via \textit{weight mirroring}, a perturbation-based algorithm for learning a credit assignment matrix \cite{akrout2019deep,lansdell2019learning}. 
In our simulations (Fig.~\ref{fig:feedback}B), the alignment between $\mathbf{M}$ and $(\mathbf{W}^{\mathrm{bmi1}})^\top$ is initially set at $0.5$, then slowly increases as $\mathbf{M}$ is learned simultaneously along with the recurrent weights $\mathbf{W}^{\mathrm{rec}}$. Instead of predicting the change in flow field before and after learning, we now apply our analysis over discrete windows during learning. As expected, activity changes are biased---and hence the SL and RL learning rules are distinguishable---early in learning. However, these differences disappear as $\mathbf{M}$ approaches $(\mathbf{W}^{\mathrm{bmi}1})^\top$ later in learning. 

\subsection{Incorporating driving feedback} 
So far, we have not included \textit{driving} feedback, which can strongly influence the dynamics of population activity. We next included fixed driving feedback weights $\mathbf{W}^{\mathrm{fb}}=\gamma \mathbf{M}$ and varied the strength of these weights by a scalar factor $\gamma$ (Fig.~\ref{fig:feedback}C). 
$\mathbf{M}$ was randomly generated such that $\mathrm{sim}(\mathbf{M},(\mathbf{W}^{\mathrm{bmi1}})^\top)=0.5$.
Note that the networks trained with SL therefore used the same mapping $\mathbf{M}$ for both the learning rule and for the driving feedback.
We found that bias in an RNN trained with SL is correctly identified for a wide range of $\gamma$ (Fig.~\ref{fig:feedback}D). 
For RNNs trained with RL, we found that the correct learning rule is identified for weak feedback, but that the FFCC metric fails for strong driving feedback (Fig.~\ref{fig:feedback}E). This is likely because neural activity in this limit is dominated by the contribution from $\mathbf{W}^{\mathrm{fb}}$, which is fully aligned with $\mathbf{M}$ and less aligned to $(\mathbf{W}^{\mathrm{bmi}1})^\top$. 

\begin{figure}[t!]

  \centering
  \includegraphics[width=1.0\textwidth]{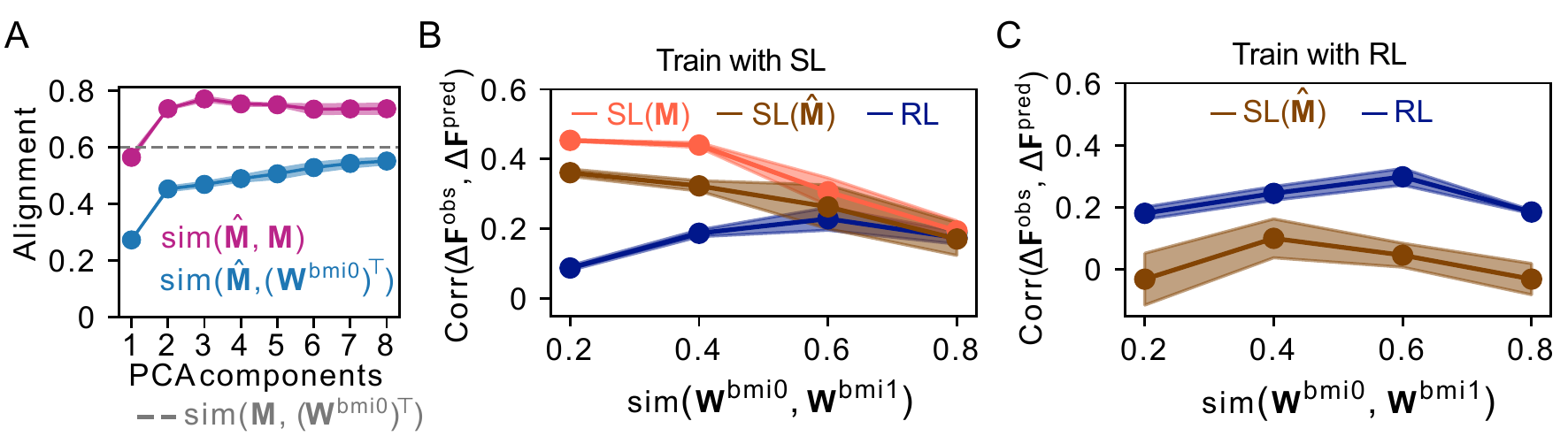}
  \caption{Estimating the credit assignment mapping.
  (A) The similarities between the estimated $\mathbf{\hat{M}}$ and the true $\mathbf{M}$ (magenta) and between $\mathbf{\hat{M}}$ and $(\mathbf{W}^\mathrm{bmi0})^\top$ (blue) as a function of the number of principal components used in the regression in an RNN trained with SL.
  (B) Comparison between predicted and observed changes in the activity flow field as a function of old vs.~new decoder alignment for an RNN trained with SL. Model predictions assume SL using the true credit assignment mapping $\mathbf{M}$ (red), SL using the $\mathbf{\hat{M}}$ estimated from pretraining (brown), or RL (blue).
  (C) Same as (B), for an RNN trained with RL (and where $\mathbf{\hat{M}}$ is estimated from pretraining).}
  \label{fig:estimate-m}
\end{figure}

\subsection{Estimating the credit assignment mapping from observed activity}\label{sec:estimating-credit-assignment}

Our approach thus far estimates the extent to which the activity updates follow either the direction of the gradient defined by the decoder weights or the (possibly biased) gradient defined by the credit assignment mapping. 
This estimation assumes knowledge of the credit assignment mapping $\mathbf{M}$.
In a neuroscience experiment, however, this quantity would not be known \textit{a priori} and would need to be estimated.
To obtain an estimate $\mathbf{\hat{M}}$ of  $\mathbf{M}$, we follow an approach previously used in experiments, in which dimensionality reduction is first applied to the neural data, followed by linear regression to relate the network activity to the cursor movements \cite{sadtler2014neural,oby2019new,feulner2021neural}. 
(A possible alternative approach could be to use the Internal Model Estimation framework proposed by \citet{golub2015internal}.)
More specifically, we (i) pretrain the RNN with decoder $\mathbf{W}^{\mathrm{bmi0}}$ and biased credit assignment mapping $\mathbf{M}$ (where $\mathrm{sim}(\mathbf{M},(\mathbf{W}^{\mathrm{bmi}0})^\top)=0.6$), (ii) observe the activity and cursor trajectories produced by the trained network, (iii) perform principal component analysis on the network activity and relate the principal components to the cursor positions with linear regression, and (iv) use the regression coefficients to define the estimated credit assignment mapping $\mathbf{\hat{M}} = \mathbf{D} \mathbf{C}$, where $\mathbf{C}$ is the matrix of principal-component eigenvectors and $\mathbf{D}$ is the matrix of regression coefficients. 
Fig.~\ref{fig:estimate-m}A shows that the $\mathbf{\hat{M}}$ obtained in this way is similar to the true $\mathbf{M}$ used to train the RNN. It is important to note for our aim of distinguishing learning rules that $\mathbf{\hat{M}}$ is more similar to $\mathbf{M}$ than to $\mathbf{W}^\mathrm{bmi0}$ and correctly identifies the alignment between $\mathbf{M}$ and $(\mathbf{W}^{\mathrm{bmi0}})^\top$. We also see from Fig.~\ref{fig:estimate-m}A that this result depends only weakly on the number of principal components used to describe the network activity.

After estimating $\mathbf{\hat{M}}$, we changed the decoder to $\mathbf{W}^{\mathrm{bmi}1}$ and trained the network to proficiency with either SL (using $\mathbf{M}$ in the learning update) or RL (which doesn't use $\mathbf{M}$). For an RNN trained with SL (Fig.~\ref{fig:estimate-m}B), the correct learning rule is identified using either the true $\mathbf{M}$ (red curve) or the estimated $\mathbf{\hat{M}}$ (brown curve) for low and intermediate degrees of similarity between $\mathbf{W}^\mathrm{bmi0}$ and $\mathbf{W}^\mathrm{bmi1}$.
For an RNN trained with RL (Fig.~\ref{fig:estimate-m}C), correct identification of the learning rule does not depend on the similarity between $\mathbf{W}^\mathrm{bmi0}$ and $\mathbf{W}^\mathrm{bmi1}$.
Together, the above results show that, for a relatively wide range of parameters, the credit assignment mapping can be estimated from observed data, and that this estimate is sufficient to distinguish between learning rules used to train different networks.

\subsection{RL with non-isotropic noise} 
In the theoretical and empirical results above, we have assumed that the noise covariance is isotropic.
If we instead allow for non-isotropic recurrent noise covariance $\boldsymbol{\Sigma}$, we find that this covariance appears in the expected RL weight update (Appendix \ref{appendix:RL}).
As in feedforward networks \cite{scott2021beyond}, this introduces a \textit{bias} in the weight updates, so that the noise-averaged weight updates no longer follow the true gradient.
Given that RL is no longer unbiased in this case, we thus asked under what conditions we might still be able to distinguish biased SL from RL.

We first established that the RNNs can learn a cursor-control task with non-isotropic noise via either SL (trivially) or RL (as long as some of the noise is in the subspace of the decoder). To do this, we pretrained networks with $\mathbf{W}^{\mathrm{bmi}0}$, changed the decoder to $\mathbf{W}^{\mathrm{bmi}1}$, and then trained using either SL or RL. 
In the SL case, we chose a biased credit mapping $\mathbf{M}$ that has partial overlap with the new decoder, with $\mathrm{sim}(\mathbf{M}, (\mathbf{W}^{\mathrm{bmi}1})^\top)=0.6$. 
The noise covariance was chosen to be $d$-dimensional, with $\mathrm{rank}(\boldsymbol{\Sigma}) = d$, and isotropic within those dimensions. The first two of these dimensions were selected to lie in the subspace spanned by $\mathbf{M}$ for the SL-trained RNNs or by $\mathbf{W}^\mathrm{bmi1}$ for the RL-trained RNNs. Other components of $\boldsymbol{\Sigma}$ were added in random dimensions orthogonal to this subspace and to one another.

\begin{figure}[t!]
  \centering
  \includegraphics[width=1.0\textwidth]{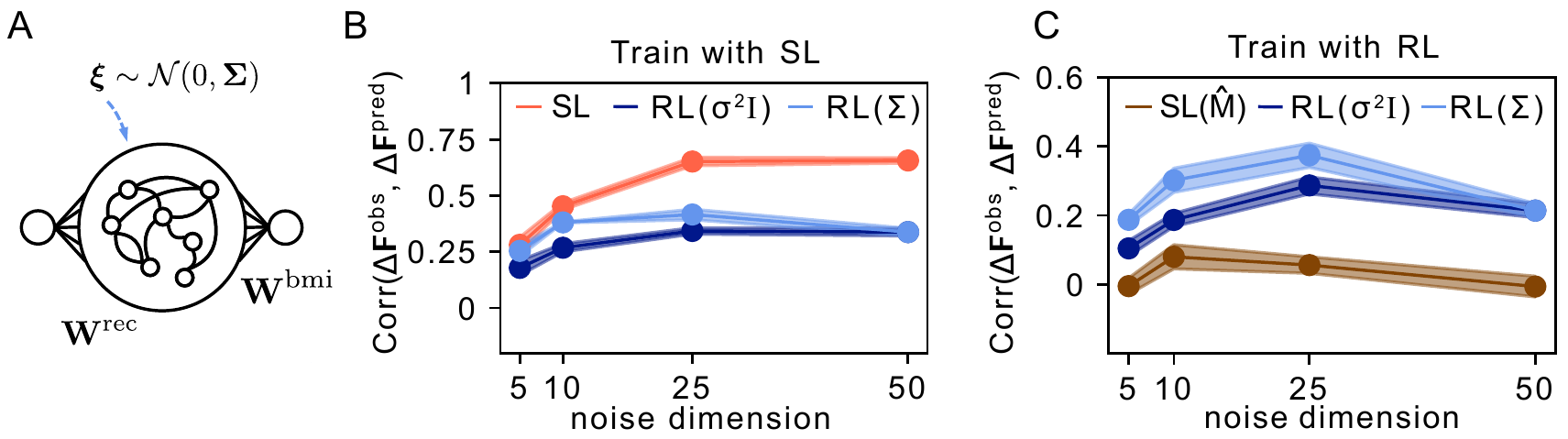}
  \caption{Distinguishing learning rules with non-isotropic noise.
  (A) Correlated noise with covariance matrix $\boldsymbol{\Sigma}$ is injected into the recurrent units of the network during training and testing.
  (B) For RNNs trained with SL, the similarities between the observed flow field change and the predicted flow field change assuming SL (red), RL using the true noise covariance (light blue), or RL using a naive isotropic estimate of the noise covariance (dark blue).
  (C) Same as (B), but for RNNs trained with RL. Brown line is correlation of observed flow field change and predicted flow field change assuming SL with random $\mathbf{\hat{M}}$, where $\mathrm{sim}(\mathbf{\hat{M}}, (\mathbf{W}^{\mathrm{bmi}1})^\top)=0.6$.}
    \label{fig:non-iso}
\end{figure}

The results in Fig.~\ref{fig:non-iso} show that, when training with SL, we cannot reliably distinguish the learning rule used to train the network when the noise is very low-dimensional  (Fig.~\ref{fig:non-iso}B). When training with RL, however, the learning rule can be distinguished even with low-dimensional noise (Fig.~\ref{fig:non-iso}C). As the noise dimensionality increases, the learning rules become more easily distinguishable in both cases. This remains true in the RL case even if we do not presume to know the true noise covariance matrix, but instead use a naive assumption of isotropic noise in order to generate the predictions (dark blue lines in Fig.~\ref{fig:non-iso}B,C).


\section{Discussion}
In this work, we have proposed a method for distinguishing biased SL from RL under the assumption that the mapping from neural activity to behavior is known, as in a BMI experiment. More generally, the approach could be used to distinguish between different candidate learning rules with different degrees or types of bias.
We have illustrated the approach by modeling a BMI cursor-control experiment in motor cortex, where there is ample experimental evidence for plasticity during motor learning \cite{peters2017learning}. Both SL and RL are plausible candidates for how motor learning might be implemented in the brain, either within motor cortex itself or within extended motor circuits involving motor cortex together with the cerebellum \cite{albus1971theory,porrill2013adaptive,raymond2018computational} or basal ganglia \cite{redgrave1999basal,athalye2020neural}. 

One possible limitation of our approach is that the number of learning rules that might conceivably be implemented in recurrent circuits in the brain is much greater than the two that we have focused on here. Nevertheless, we view the SL and RL rules considered above as representative examples of two fundamental classes of learning rules: one in which learning is based on a model for credit assignment, where this model is necessarily imperfect, and another which uses no such model. While we cannot test every learning algorithm from these classes, we conjecture that the basic fact that allows us to distinguish our SL and RL rules---namely, that the expected flow field change depends on a credit-assignment matrix for model-based rules but not for non-model-based rules---is a fairly generic feature of these two classes. To support this, we show in Appendix \ref{sec:biased_bptt} that our main results hold for a different version of the biased learning rule that we have studied.


While we have shown that different learning rules are in principle distinguishable in simulation, there will be challenges in extending this analysis to real data. 
Reliable estimation of the FFCC metric might require large populations of neurons recorded for long periods of time. Other potential challenges that will need to be addressed include accounting for the possibility that multiple types of learning might be occurring simultaneously at different locations, as well as considering cognitive mechanisms such as attention and contextual inference, which have not been included in this model. Finally, there are possible ambiguities to consider in mapping our model onto the brain's motor circuitry. While the most straightforward interpretation is that the RNN in our model corresponds to motor cortex, a more elaborate interpretation could be made by noting that both the cerebellum and the basal ganglia have well-established roles in motor learning \cite{krakauer2019motor} and are likely to play roles in BMI learning as well \cite{athalye2020neural,hennig2021howlearning}.

\begin{ack}
The authors would like to acknowledge Larry Abbott, Andrew Zimnik, Laureline Logiaco, Kaushik Lakshminarasimhan, and Vivek Athalye for fruitful discussions. Thanks to Amin Nejatbakhsh, David Clark, and Danil Tyulmankov for feedback on the manuscript.
Support for this work was provided by NIH-NINDS (R00NS114194).
\end{ack}

\newpage


\bibliographystyle{unsrtnat} 
\bibliography{references}

\begin{thebibliography}{40}
\providecommand{\natexlab}[1]{#1}
\providecommand{\url}[1]{\texttt{#1}}
\expandafter\ifx\csname urlstyle\endcsname\relax
  \providecommand{\doi}[1]{doi: #1}\else
  \providecommand{\doi}{doi: \begingroup \urlstyle{rm}\Url}\fi

\bibitem[Grossberg(1987)]{grossberg1987competitive}
Stephen Grossberg.
\newblock Competitive learning: From interactive activation to adaptive
  resonance.
\newblock \emph{Cognitive Science}, 11\penalty0 (1):\penalty0 23--63, 1987.

\bibitem[Richards et~al.(2019)Richards, Lillicrap, Beaudoin, Bengio, Bogacz,
  Christensen, Clopath, Costa, de~Berker, Ganguli, et~al.]{richards2019deep}
Blake~A Richards, Timothy~P Lillicrap, Philippe Beaudoin, Yoshua Bengio, Rafal
  Bogacz, Amelia Christensen, Claudia Clopath, Rui~Ponte Costa, Archy
  de~Berker, Surya Ganguli, et~al.
\newblock A deep learning framework for neuroscience.
\newblock \emph{Nature Neuroscience}, 22\penalty0 (11):\penalty0 1761--1770,
  2019.

\bibitem[Lillicrap et~al.(2020)Lillicrap, Santoro, Marris, Akerman, and
  Hinton]{lillicrap2020backpropagation}
Timothy~P Lillicrap, Adam Santoro, Luke Marris, Colin~J Akerman, and Geoffrey
  Hinton.
\newblock Backpropagation and the brain.
\newblock \emph{Nature Reviews Neuroscience}, 21\penalty0 (6):\penalty0
  335--346, 2020.

\bibitem[Lillicrap et~al.(2016)Lillicrap, Cownden, Tweed, and
  Akerman]{lillicrap2016random}
Timothy~P Lillicrap, Daniel Cownden, Douglas~B Tweed, and Colin~J Akerman.
\newblock Random synaptic feedback weights support error backpropagation for
  deep learning.
\newblock \emph{Nature Communications}, 7\penalty0 (1):\penalty0 1--10, 2016.

\bibitem[Murray(2019)]{murray2019local}
James~M Murray.
\newblock Local online learning in recurrent networks with random feedback.
\newblock \emph{eLife}, 8:\penalty0 e43299, 2019.

\bibitem[Akrout et~al.(2019)Akrout, Wilson, Humphreys, Lillicrap, and
  Tweed]{akrout2019deep}
Mohamed Akrout, Collin Wilson, Peter Humphreys, Timothy Lillicrap, and
  Douglas~B Tweed.
\newblock Deep learning without weight transport.
\newblock \emph{Advances in Neural Information Processing Systems}, 32, 2019.

\bibitem[Amit(2019)]{amit2019deep}
Yali Amit.
\newblock Deep learning with asymmetric connections and hebbian updates.
\newblock \emph{Frontiers in Computational Neuroscience}, 13:\penalty0 18,
  2019.

\bibitem[Williams(1992)]{williams1992simple}
Ronald~J Williams.
\newblock Simple statistical gradient-following algorithms for connectionist
  reinforcement learning.
\newblock \emph{Machine Learning}, 8\penalty0 (3-4):\penalty0 229--256, 1992.

\bibitem[Werfel et~al.(2003)Werfel, Xie, and Seung]{werfel2003learning}
Justin Werfel, Xiaohui Xie, and H~Seung.
\newblock Learning curves for stochastic gradient descent in linear feedforward
  networks.
\newblock \emph{Advances in Neural Information Processing Systems}, 16, 2003.

\bibitem[Miconi(2017)]{miconi2017biologically}
Thomas Miconi.
\newblock Biologically plausible learning in recurrent neural networks
  reproduces neural dynamics observed during cognitive tasks.
\newblock \emph{eLife}, 6:\penalty0 e20899, 2017.

\bibitem[Sutton and Barto(2018)]{sutton2018reinforcement}
Richard~S Sutton and Andrew~G Barto.
\newblock \emph{Reinforcement Learning: An Introduction}.
\newblock MIT Press, 2018.

\bibitem[Sadtler et~al.(2014)Sadtler, Quick, Golub, Chase, Ryu, Tyler-Kabara,
  Byron, and Batista]{sadtler2014neural}
Patrick~T Sadtler, Kristin~M Quick, Matthew~D Golub, Steven~M Chase, Stephen~I
  Ryu, Elizabeth~C Tyler-Kabara, M~Yu Byron, and Aaron~P Batista.
\newblock Neural constraints on learning.
\newblock \emph{Nature}, 512\penalty0 (7515):\penalty0 423--426, 2014.

\bibitem[Zhou et~al.(2019)Zhou, Tien, Ravikumar, and Chase]{zhou2019distinct}
Xiao Zhou, Rex~N Tien, Sadhana Ravikumar, and Steven~M Chase.
\newblock Distinct types of neural reorganization during long-term learning.
\newblock \emph{Journal of Neurophysiology}, 121\penalty0 (4):\penalty0
  1329--1341, 2019.

\bibitem[Losey et~al.(2022)Losey, Hennig, Oby, Golub, Sadlter, Quick, Ryu,
  Tyler-Kabara, Batista, Byron, et~al.]{losey2022learning}
Darby~M Losey, Jay~A Hennig, Emily~R Oby, Matthew~D Golub, Patrick~T Sadlter,
  Kristin~M Quick, Stephen~I Ryu, Elizabeth~C Tyler-Kabara, Aaron~P Batista,
  M~Yu Byron, et~al.
\newblock Learning alters neural activity to simultaneously support memory and
  action.
\newblock \emph{bioRxiv}, 2022.

\bibitem[Athalye et~al.(2017)Athalye, Ganguly, Costa, and
  Carmena]{athalye2017emergence}
Vivek~R Athalye, Karunesh Ganguly, Rui~M Costa, and Jose~M Carmena.
\newblock Emergence of coordinated neural dynamics underlies neuroprosthetic
  learning and skillful control.
\newblock \emph{Neuron}, 93\penalty0 (4):\penalty0 955--970, 2017.

\bibitem[Oby et~al.(2019)Oby, Golub, Hennig, Degenhart, Tyler-Kabara, Byron,
  Chase, and Batista]{oby2019new}
Emily~R Oby, Matthew~D Golub, Jay~A Hennig, Alan~D Degenhart, Elizabeth~C
  Tyler-Kabara, M~Yu Byron, Steven~M Chase, and Aaron~P Batista.
\newblock New neural activity patterns emerge with long-term learning.
\newblock \emph{Proceedings of the National Academy of Sciences}, 116\penalty0
  (30):\penalty0 15210--15215, 2019.

\bibitem[Sorrell et~al.(2021)Sorrell, Rule, and O'Leary]{sorrell2021brain}
Ethan Sorrell, Michael~E Rule, and Timothy O'Leary.
\newblock Brain--machine interfaces: Closed-loop control in an adaptive system.
\newblock \emph{Annual Review of Control, Robotics, and Autonomous Systems},
  4:\penalty0 167--189, 2021.

\bibitem[Lansdell et~al.(2019)Lansdell, Prakash, and
  Kording]{lansdell2019learning}
Benjamin~James Lansdell, Prashanth~Ravi Prakash, and Konrad~Paul Kording.
\newblock Learning to solve the credit assignment problem.
\newblock \emph{International Conference on Learning Representations}, 2019.

\bibitem[Lim et~al.(2015)Lim, McKee, Woloszyn, Amit, Freedman, Sheinberg, and
  Brunel]{lim2015inferring}
Sukbin Lim, Jillian~L McKee, Luke Woloszyn, Yali Amit, David~J Freedman,
  David~L Sheinberg, and Nicolas Brunel.
\newblock Inferring learning rules from distributions of firing rates in
  cortical neurons.
\newblock \emph{Nature Neuroscience}, 18\penalty0 (12):\penalty0 1804--1810,
  2015.

\bibitem[Nayebi et~al.(2020)Nayebi, Srivastava, Ganguli, and
  Yamins]{nayebi2020identifying}
Aran Nayebi, Sanjana Srivastava, Surya Ganguli, and Daniel~L Yamins.
\newblock Identifying learning rules from neural network observables.
\newblock \emph{Advances in Neural Information Processing Systems}, 33, 2020.

\bibitem[Kepple et~al.(2021)Kepple, Engelken, and Rajan]{kepple2021curriculum}
Daniel~R Kepple, Rainer Engelken, and Kanaka Rajan.
\newblock Curriculum learning as a tool to uncover learning principles in the
  brain.
\newblock \emph{International Conference on Learning Representations}, 2021.

\bibitem[Feulner and Clopath(2021)]{feulner2021neural}
Barbara Feulner and Claudia Clopath.
\newblock Neural manifold under plasticity in a goal driven learning behaviour.
\newblock \emph{PLoS Computational Biology}, 17\penalty0 (2):\penalty0
  e1008621, 2021.

\bibitem[Feulner et~al.(2022)Feulner, Perich, Chowdhury, Miller, Gallego, and
  Clopath]{feulner2022small}
Barbara Feulner, Matthew~G Perich, Raeed~H Chowdhury, Lee~E Miller, Juan~A
  Gallego, and Claudia Clopath.
\newblock Small, correlated changes in synaptic connectivity may facilitate
  rapid motor learning.
\newblock \emph{Nature Communications}, 13\penalty0 (1):\penalty0 1--14, 2022.

\bibitem[Golub et~al.(2018)Golub, Sadtler, Oby, Quick, Ryu, Tyler-Kabara,
  Batista, Chase, and Byron]{golub2018learning}
Matthew~D Golub, Patrick~T Sadtler, Emily~R Oby, Kristin~M Quick, Stephen~I
  Ryu, Elizabeth~C Tyler-Kabara, Aaron~P Batista, Steven~M Chase, and M~Yu
  Byron.
\newblock Learning by neural reassociation.
\newblock \emph{Nature Neuroscience}, 21\penalty0 (4):\penalty0 607--616, 2018.

\bibitem[Rumelhart et~al.(1985)Rumelhart, Hinton, and
  Williams]{rumelhart1985learning}
David~E Rumelhart, Geoffrey~E Hinton, and Ronald~J Williams.
\newblock Learning internal representations by error propagation.
\newblock Technical report, California Univ San Diego La Jolla Inst for
  Cognitive Science, 1985.

\bibitem[Williams and Zipser(1989)]{williams1989learning}
Ronald~J Williams and David Zipser.
\newblock A learning algorithm for continually running fully recurrent neural
  networks.
\newblock \emph{Neural Computation}, 1\penalty0 (2):\penalty0 270--280, 1989.

\bibitem[Marbach and Tsitsiklis(2001)]{marbach2001simulation}
Peter Marbach and John~N Tsitsiklis.
\newblock Simulation-based optimization of markov reward processes.
\newblock \emph{IEEE Transactions on Automatic Control}, 46\penalty0
  (2):\penalty0 191--209, 2001.

\bibitem[Fiete and Seung(2006)]{fiete2006gradient}
Ila~R Fiete and H~Sebastian Seung.
\newblock Gradient learning in spiking neural networks by dynamic perturbation
  of conductances.
\newblock \emph{Physical Review Letters}, 97\penalty0 (4):\penalty0 048104,
  2006.

\bibitem[Golub et~al.(2015)Golub, Byron, and Chase]{golub2015internal}
Matthew~D Golub, M~Yu Byron, and Steven~M Chase.
\newblock Internal models for interpreting neural population activity during
  sensorimotor control.
\newblock \emph{eLife}, 4:\penalty0 e10015, 2015.

\bibitem[Scott and Frank(2021)]{scott2021beyond}
Daniel~N. Scott and Michael~J. Frank.
\newblock Beyond gradients: Noise correlations control hebbian plasticity to
  shape credit assignment.
\newblock \emph{bioRxiv}, 2021.

\bibitem[Peters et~al.(2017)Peters, Liu, and Komiyama]{peters2017learning}
Andrew~J Peters, Haixin Liu, and Takaki Komiyama.
\newblock Learning in the rodent motor cortex.
\newblock \emph{Annual Review of Neuroscience}, 40:\penalty0 77, 2017.

\bibitem[Albus(1971)]{albus1971theory}
James~S Albus.
\newblock A theory of cerebellar function.
\newblock \emph{Mathematical Biosciences}, 10\penalty0 (1-2):\penalty0 25--61,
  1971.

\bibitem[Porrill et~al.(2013)Porrill, Dean, and Anderson]{porrill2013adaptive}
John Porrill, Paul Dean, and Sean~R Anderson.
\newblock Adaptive filters and internal models: multilevel description of
  cerebellar function.
\newblock \emph{Neural Networks}, 47:\penalty0 134--149, 2013.

\bibitem[Raymond and Medina(2018)]{raymond2018computational}
Jennifer~L Raymond and Javier~F Medina.
\newblock Computational principles of supervised learning in the cerebellum.
\newblock \emph{Annual Review of Neuroscience}, 41:\penalty0 233--253, 2018.

\bibitem[Redgrave et~al.(1999)Redgrave, Prescott, and
  Gurney]{redgrave1999basal}
Peter Redgrave, Tony~J Prescott, and Kevin Gurney.
\newblock The basal ganglia: a vertebrate solution to the selection problem?
\newblock \emph{Neuroscience}, 89\penalty0 (4):\penalty0 1009--1023, 1999.

\bibitem[Athalye et~al.(2020)Athalye, Carmena, and Costa]{athalye2020neural}
Vivek~R Athalye, Jose~M Carmena, and Rui~M Costa.
\newblock Neural reinforcement: re-entering and refining neural dynamics
  leading to desirable outcomes.
\newblock \emph{Current Opinion in Neurobiology}, 60:\penalty0 145--154, 2020.

\bibitem[Krakauer et~al.(2019)Krakauer, Hadjiosif, Xu, Wong, and
  Haith]{krakauer2019motor}
John~W Krakauer, Alkis~M Hadjiosif, Jing Xu, Aaron~L Wong, and Adrian~M Haith.
\newblock Motor learning.
\newblock \emph{Comprehensive Physiology}, 9\penalty0 (2):\penalty0 613--663,
  2019.

\bibitem[Hennig et~al.(2021)Hennig, Oby, Losey, Batista, Byron, and
  Chase]{hennig2021howlearning}
Jay~A Hennig, Emily~R Oby, Darby~M Losey, Aaron~P Batista, M~Yu Byron, and
  Steven~M Chase.
\newblock How learning unfolds in the brain: toward an optimization view.
\newblock \emph{Neuron}, 2021.

\bibitem[Gerstner et~al.(2018)Gerstner, Lehmann, Liakoni, Corneil, and
  Brea]{gerstner2018eligibility}
Wulfram Gerstner, Marco Lehmann, Vasiliki Liakoni, Dane Corneil, and Johanni
  Brea.
\newblock Eligibility traces and plasticity on behavioral time scales:
  experimental support of neohebbian three-factor learning rules.
\newblock \emph{Frontiers in Neural Circuits}, 12:\penalty0 53, 2018.

\bibitem[Marschall et~al.(2020)Marschall, Cho, and Savin]{marschall2020unified}
Owen Marschall, Kyunghyun Cho, and Cristina Savin.
\newblock A unified framework of online learning algorithms for training
  recurrent neural networks.
\newblock \emph{Journal of Machine Learning Research}, 2020.

\end{thebibliography}

\newpage



\newpage
\beginsupplement 

\appendix


\section{Distinguishing supervised learning from reinforcement learning in a feedforward model}
\label{appendix:feedforward}
\begin{figure}[t!]
  \centering
  \includegraphics[width=0.9\textwidth]{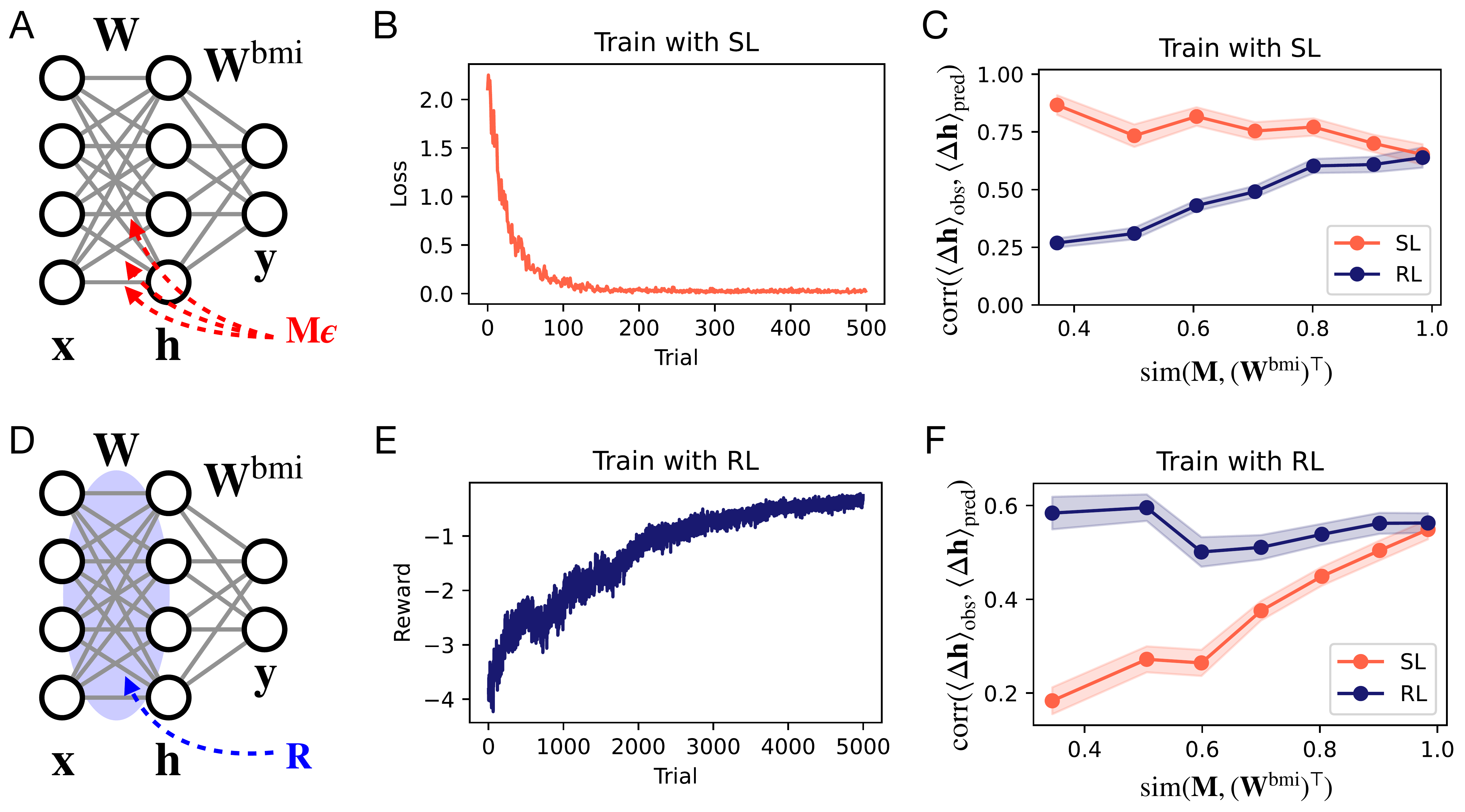}
  \caption{Inference of learning rules from activity in a feedforward model. (A) Hidden-layer weights in a linear network are trained with SL to map random input patterns onto random output patterns. 
  (B) An example showing that the loss successfully decreases during training.
  (C) Comparison of the observed change in hidden-layer activity during training with the predicted change assuming SL (red) vs.~assuming RL (blue).
  (D-F) Same as top row, but for a network trained with RL.}
 \label{fig-feedforward}
\end{figure}

In order to illustrate the main idea from our paper in a simplified context, we show in this section how observed hidden-layer activity in a linear feedforward network can be used to infer the learning rule that is used to train the network.
Consider the simple feedforward network shown in Fig.~\ref{fig-feedforward}. In this network, random binary input patterns $\mathbf{x}^t$, where $x_i^t \in \{-1,1\}$ and $t = 1, \ldots, T$, are projected onto a hidden layer $\mathbf{h}^t = \mathbf{W}\mathbf{x}^t + \boldsymbol{\xi}^t$, where $\boldsymbol{\xi}^t \sim \mathcal{N}(0,\boldsymbol{\Sigma})$ is noise injected into the network. The readout is then given by $\mathbf{y}^t = \mathbf{W}^\mathrm{bmi} \mathbf{h}$, and the goal of the network is to minimize $L = \sum_t L_t = \frac{1}{T}\sum_t |\mathbf{y}^{* t} - \mathbf{y}^t|^2$, where $\mathbf{y}^*$ are randomly chosen target patterns with $y_i^* \sim \mathcal{N}(0, 1)$. 

With $\mathbf{W}^\mathrm{bmi}$ fixed, we can train the network using either supervised learning (SL) or reinforcement learning (RL). In the case of SL, the learning rule performs credit assignment using a model $\mathbf{M}$ that approximates the ideal credit assignment model $(\mathbf{W}^\mathrm{bmi})^\top$ to project the error $\boldsymbol{\epsilon}^t = \mathbf{y}^{*t} - \mathbf{y}^t$ back to the hidden layer (Fig.~\ref{fig-feedforward}A), giving the following weight update:
\begin{equation}\label{eq:feedforwardSL}
    \Delta W_{ij}^\mathrm{SL} = \eta_\mathrm{SL} \sum_t [\mathbf{M} \boldsymbol{\epsilon}^t]_i x_j^t.
\end{equation}
In the case where $\mathbf{M} = (\mathbf{W} ^\mathrm{bmi})^\top$, this rule would be implementing gradient descent. In the case where $\mathbf{M} \neq (\mathbf{W} ^\mathrm{bmi})^\top$ but the two matrices have positive alignment, it instead implements a biased version of gradient descent.
This is similar to learning with Feedback Alignment \cite{lillicrap2016random}, except that here we do not assume that the readout weights are being learned.

An alternative learning algorithm is policy gradient learning \cite{williams1992simple}, which gives the following update equation:
\begin{equation}\label{eq:feedforwardRL}
    \Delta W_{ij}^\mathrm{RL} = \eta_\mathrm{RL} \sum_t (R^t - \bar{R}^t) \xi_i^t x_j^t,
\end{equation}
where we define the reward as $R^t = -L_t$, and we subtract off the baseline $\bar{R}^t = \langle L_t \rangle_\xi$.

Our goal will be to train the network to minimize the loss using either the SL or RL learning rule, then, assuming $\mathbf{W}^\mathrm{bmi}$ and $\mathbf{M}$ are known, to use the observed output and hidden-layer activity during training to infer which of the two algorithms was used to train the network. 
In the case of SL, averaging \eqref{eq:feedforwardSL} over noise $\xi$ gives
\begin{equation}\label{eq:feedforwardSL_dW}
    \langle\Delta W_{ij}^\mathrm{SL}\rangle = \eta_\mathrm{SL} \sum_t [\mathbf{M} \langle\boldsymbol{\epsilon}^t\rangle]_i x_j^t.
\end{equation}
The expected change in the hidden-layer activity due to learning is then
\begin{equation}\label{eq:feedforwardSL_dh}
\begin{split}
    \langle \Delta\mathbf{h}^t \rangle 
    &= \langle\Delta \mathbf{W}^\mathrm{SL}\rangle \mathbf{x}^t \\
    &= \eta_\mathrm{SL} \sum_{t'}
    (\mathbf{x}^t \cdot \mathbf{x}^{t'})
    \mathbf{M} \langle\boldsymbol{\epsilon}^{t'}\rangle .
\end{split}
\end{equation}
In the case of RL, the average of the weight update from \eqref{eq:feedforwardRL} is 
\begin{equation}\label{eq:feedforwardRL_dW}
    \langle\Delta W_{ij}^\mathrm{RL}\rangle = \eta_\mathrm{RL} \sum_t [\boldsymbol{\Sigma}(\mathbf{W}^\mathrm{bmi})^\top \langle\boldsymbol{\epsilon}^t\rangle]_i x_j^t.
\end{equation}
The expected change in the hidden-layer activity in this case is then
\begin{equation}\label{eq:feedforwardRL_dh}
\begin{split}
    \langle \Delta\mathbf{h}^t \rangle 
    &= \langle\Delta \mathbf{W}^\mathrm{RL}\rangle \mathbf{x}^t \\
    &= \eta_\mathrm{RL} \sum_{t'} (\mathbf{x}^t \cdot \mathbf{x}^{t'}) \boldsymbol{\Sigma} (\mathbf{W}^\mathrm{bmi})^\top \langle\boldsymbol{\epsilon}^{t'}\rangle .
\end{split}
\end{equation}

Equations \eqref{eq:feedforwardSL_dh} and \eqref{eq:feedforwardRL_dh} provide predictions for how the hidden-layer activity is expected to evolve under either SL or RL. Since we do not presume to have knowledge of either the learning rate or the upstream activity $\mathbf{x}^t$, we can concern ourselves only with the direction, but not the magnitude, of $\langle \Delta \mathbf{h}^t \rangle$. In addition, we must make an assumption about correlations between the input patterns. In general, if the input patterns are correlated with one another, we can let $\mathbf{x}^t \cdot \mathbf{x}^{t'} = C(|t - t'|)$, where $C$ is a function with a peak near zero. 

For the simulations shown in Fig.~\ref{fig-feedforward}, we have used uncorrelated inputs $\mathbf{x}^t$, so we can assume that $\mathbf{x}^t \cdot \mathbf{x}^{t'} \sim \delta_{tt'}$. With this assumption, Equations \eqref{eq:feedforwardSL_dh} and \eqref{eq:feedforwardRL_dh} give
\begin{equation}\label{eq:feedforward_dh_pred}
\begin{split}
    \langle \Delta\mathbf{h}^t \rangle_\mathrm{pred} \propto 
    \begin{cases}
    \mathbf{M} \langle \boldsymbol{\epsilon}^t \rangle,
    \quad\quad & \mathrm{SL} \\
    \boldsymbol{\Sigma} (\mathbf{W}^\mathrm{bmi})^\top \langle \boldsymbol{\epsilon}^t \rangle.
    \quad\quad & \mathrm{RL}
    \end{cases}
\end{split}
\end{equation}
Thus, if we assume that the quantities $\mathbf{M}$, $\boldsymbol{\Sigma}$, and $(\mathbf{W}^\mathrm{bmi})^\top$ are known or can be estimated, \eqref{eq:feedforward_dh_pred} provides a prediction for how the hidden-layer activity is expected to change due to learning. If this quantity is estimated from data, then the average over noise can be replaced with an empirical average over observed trials: $\langle \ldots \rangle = \frac{1}{N_\mathrm{trials}}\sum_{n =1}^{N_\mathrm{trials}} (\ldots)$.  

If we next suppose that we observe the hidden-layer activity empirically without necessarily knowing the learning rule being used to train the network, then we can define the observed change in hidden-layer activity as the difference between activity observed in early and late trials
\begin{equation}\label{eq:feedforward_dh_obs}
    \langle \Delta\mathbf{h}^t \rangle_\mathrm{obs} = 
    \frac{1}{N_\mathrm{late}} \sum_{n=1}^{N_\mathrm{late}} \mathbf{h}^{n,t}
    - \frac{1}{N_\mathrm{early}} \sum_{n=1}^{N_\mathrm{early}} \mathbf{h}^{n,t},
\end{equation}
where $\mathbf{h}^{n,t}$ is the activity observed at time $t$ in trial $n$. We can then compare the similarity of $ \langle \Delta\mathbf{h}^t \rangle_\mathrm{obs}$ with $\langle \Delta\mathbf{h}^t \rangle_\mathrm{pred}$ by computing their correlations: 
\begin{equation}
\label{eq:feedforward_corr}
\mathrm{corr}(\langle \Delta\mathbf{h}^t \rangle_\mathrm{obs}, \langle \Delta\mathbf{h}^t \rangle_\mathrm{pred}) = \frac{ \sum_i 
(\langle \Delta h_i \rangle_\mathrm{obs} - \overline{\langle \Delta h \rangle_\mathrm{obs}})
(\langle \Delta h_i \rangle_\mathrm{pred} - \overline{\langle \Delta h \rangle_\mathrm{pred}})}
{\mathrm{std}(\langle \Delta\mathbf{h}^t \rangle_\mathrm{obs})
~ \mathrm{std}(\langle \Delta\mathbf{h}^t \rangle_\mathrm{pred})},
\end{equation}
where $\overline{(\ldots)}$ denotes an average over neurons.
In Fig.~\ref{fig-feedforward}, we show results from a linear feedforward network trained with either SL (Fig.~\ref{fig-feedforward}A-B) or RL (Fig.~\ref{fig-feedforward}D-E) to map random input patterns onto random output patterns. Then, using  \eqref{eq:feedforward_corr}, we ask whether the change in the hidden-layer activity during learning in each of these cases is more similar to $\langle \Delta \mathbf{h} \rangle_\mathrm{pred}$ predicted assuming SL or RL. The results in Fig.~\ref{fig-feedforward}C,F show that, whenever $\mathbf{M}$ and $(\mathbf{W}^\mathrm{bmi})^\top$ are sufficiently different, this metric is able to correctly identify whether the network was trained with SL (Fig.~\ref{fig-feedforward}C) or RL (Fig.~\ref{fig-feedforward}F).

\section{Biologically plausible learning rules for recurrent neural networks}
In this section we provide derivations of the two learning rules studied in our paper. 
The RNN update equation is
\begin{equation}\label{eq:rnn_update}
    \mathbf{h}^t = \left(1-\frac{1}{\tau}\right) \mathbf{h}^{t-1} + \frac{1}{\tau}\phi(\mathbf{u}^t) + \bm{\xi}^t,
\end{equation}
where $\mathbf{u}^t = \mathbf{W}^\mathrm{rec} \mathbf{h}^{t-1} + \mathbf{W}^\mathrm{fb} \mathbf{y}^{t-1} + \mathbf{W}^\mathrm{in} \mathbf{x}^t$, and $\xi_i^t \sim \mathcal{N}(0,\sigma^2_{\mathrm{rec}})$ is i.i.d.~noise injected into the network. The readout is given by $\mathbf{y} = \mathbf{W}^\mathrm{bmi} \mathbf{h}^t$. The goal of both learning rules is to iteratively update the recurrent weights $\mathbf{W}^\mathrm{rec}$ such that the readout matches a target function $\mathbf{y}^{* t}$, minimizing the magnitude of the error $\bm{\epsilon}^t = \mathbf{y}^{* t} - \mathbf{y}^t$.
The learning rules that we consider below depend on a multiplicative combination of pre- and postsynaptic activity, as well as a third factor related to error or reward. Evidence for such ``three-factor" learning rules has been found in a number of neuroscience experiments \cite{gerstner2018eligibility}.


\subsection{Random Feedback Local Online (RFLO) learning}
\label{appendix:SL}
In this section, we briefly recapitulate the derivation of RFLO \citep{murray2019local}, a supervised learning algorithm for RNNs that uses local weight updates to approximate gradient descent. 
We then compute the expectation of the weight update by averaging over noise and show that this expected weight update is determined by the credit-assignment matrix $\mathbf{M}$. 

The loss function to be minimized is
\begin{equation}
L = \frac{1}{2T} \sum_{t=1}^T \sum_{k=1}^{N_y} \Big[ y_k^{* t} - y_k^t \Big]^2.
\end{equation}
Taking the derivative with respect to the recurrent weights gives
\begin{equation}
\frac{\partial L}{\partial W_{ab}} = -\frac{1}{T} \sum_{t=1}^T \sum_{i} \big[  (\mathbf{W}^{\text{bmi}})^{\top} \bm{\epsilon}^t  \big]_i \frac{\partial h_i^t}{\partial W_{ab}}.
\end{equation}
Using (\ref{eq:rnn_update}), we obtain the following recursion relation:
\begin{equation}\label{eq:recursion}
\begin{split}
    \frac{\partial h_i^t}{\partial W^\mathrm{rec}_{ab}}
    &= \left(1 - \frac{1}{\tau} \right)
    \frac{\partial h_i^{t-1}}{\partial W^\mathrm{rec}_{ab}} 
    + \frac{1}{\tau} \delta_{ia} \phi'(u_i^t) h_b^{t-1} \\
    & \quad\quad + \frac{1}{\tau} \phi'\left( u_i^t \right)
    \left[ \sum_j W^\mathrm{rec}_{ij} \frac{\partial h_j^{t-1}}{\partial W^\mathrm{rec}_{ab}}
    + \sum_j W^\mathrm{fb}_{ij} \frac{\partial y_j^{t-1}}{\partial W^\mathrm{rec}_{ab}}\right].
\end{split}
\end{equation}
To obtain a learning rule for $W^\mathrm{rec}_{ab}$ that is local, i.e.~that depends only on pre- and postsynaptic activity from units $b$ and $a$, respectively, we can discard the second line of this equation and write $\partial h_i^t/\partial W^\mathrm{rec}_{ab} \approx \delta_{ia} p_{ab}^t$, where the eligibility trace $p_{ab}^t$ follows the recursion relation
\begin{equation}\label{eq:recursion_approx}
    p_{ab}^t 
    = \left(1 - \frac{1}{\tau} \right)
    p_{ab}^{t-1}
    + \frac{1}{\tau} \phi'\left( u_a^{t-1} \right)
    h_b^{t-1},
\end{equation}
and $p_{ab}^0 = 0$.
With this approximation, we arrive at the RFLO learning rule:
\begin{equation}\label{eq:dW_rflo}
    \Delta W^\mathrm{rec}_{ab} = \eta \sum_t [\mathbf{M} \bm{\epsilon}^t]_a p_{ab}^t.
\end{equation}

In order to compute the expected change in the RNN flow fields, we wish to compute the expectation of the weight update \eqref{eq:dW_rflo} by averaging over noise. In order to obtain a result for $\Delta \mathbf{F}$ that does not depend explicitly on $\mathbf{W}^\mathrm{rec}$, we assume a linear network with $\phi(u) = u$. For convenience, we also switch to the continuous-time limit. In this case, the above eligibility trace in \eqref{eq:recursion_approx} is given by
\begin{equation}\label{eq:p_linear}
    p_{ab} = \int_0^t e^{(s-t)/\tau} h^s_b \frac{ds}{\tau} .
\end{equation}
Thus, the expected weight update is given by
\begin{equation}\label{eq:dW_avg_SL}
\begin{split}
\langle \Delta W^\mathrm{rec}_{ij} \rangle &= \eta \sum_t\sum_k M_{ik} \int_0^t e^{(s-t)/\tau} \left\langle (y^*_k - y^t_k) h^s_j\right\rangle \frac{ds}{\tau}\\
&=\eta \sum_t\sum_k M_{ik} \int_0^t e^{(s-t)/\tau} \left(\langle \epsilon^t_k \rangle \langle h^s_j \rangle - \sum_l W^{\text{bmi}}_{kl} Q^{t,s}_{lj} \right)\frac{ds}{\tau},
\end{split}
\end{equation}
where we have defined the covariance matrix $\mathbf{Q}^{t,t'} = \langle \mathbf{h}^t (\mathbf{h}^{t'})^\top \rangle - \langle \mathbf{h}^t \rangle\, \langle \mathbf{h}^{t'} \rangle^\top$.
In the case where $\tau$ is unknown (as it may be in experimental data) or sufficiently small relative to the timescale of RNN dynamics, this expression can be simplified by taking the limit $\tau \to 0$, leading to
\begin{equation}
\label{eq:dW_av_rflo}
\langle \Delta W^\mathrm{rec}_{ij} \rangle 
= \eta \sum_t\sum_{k} M_{ik} \left( \langle \epsilon^t_k \rangle \langle h^t_j \rangle - \sum_r W^{\text{bmi}}_{kr} Q^{t,t}_{rj}\right) .
\end{equation}
In the limit where the noise is small, the first term in parentheses will be much larger than the second, which can be dropped to obtain a simplified expression.

\subsection{Reinforcement learning in recurrent neural networks}
\label{appendix:RL}
In this section, we derive a local RNN update rule using policy gradient learning. 
The resulting ``node perturbation'' learning algorithm is essentially equivalent to previously proposed RL rules for recurrent circuits from Refs.~\cite{fiete2006gradient,miconi2017biologically}.
We then compute the expectation of the weight update \eqref{eq:dW_avg_RL} by averaging over noise and show that this expected weight update is determined by the decoder $\mathbf{W}^{\mathrm{bmi}}$, following the unbiased gradient direction.


In policy gradient learning, a policy $\pi(\mathrm{action} | \mathrm{state})$ is optimized with respect to its parameters in order to maximize a scalar performance measure $R^t$. In our case, we interpret $\mathbf{h}^t$ as the action, $\mathbf{h}^{t-1}$ as the state, and $\mathbf{W}^\mathrm{rec}$ as the parameters to be optimized.
We take the policy to be 
\begin{equation}
    \pi(\mathbf{h}^t | \mathbf{h}^{t-1}, \mathbf{W}^\mathrm{rec}) \sim \mathcal{N} (\mathbf{h}^t | \boldsymbol{\mu}^t, \sigma^2_{\mathrm{rec}} \mathbf{I}),  
\end{equation}
where
\begin{equation}\label{eq:mu}
    \bm{\mu}(t, \mathbf{W}^\mathrm{rec}) = \left(1-\frac{1}{\tau}\right) \mathbf{h}^{t-1} + \frac{1}{\tau}\phi(\mathbf{u}^t)
\end{equation}
is the deterministic part of the update equation (\ref{eq:rnn_update}). 
The policy gradient theorem allows us to update the policy parameters in a way that ensures improvement of the objective $R^t$. The REINFORCE algorithm \cite{williams1992simple} is based on the policy gradient theorem and updates the parameters $\mathbf{W}^\mathrm{rec}$ according to
\begin{equation}\label{eq:rl_update0}
\Delta \mathbf{W}^\mathrm{rec} \propto \left(R^t - \bar{R}^t \right) \nabla \ln \pi(\mathbf{h}^t | \mathbf{h}^{t-1},\mathbf{W}^\mathrm{rec}) ,
\end{equation}
where the gradient is with respect to $\mathbf{W}^\mathrm{rec}$. The reward baseline $\bar{R}^t$, to be defined below, is not required for policy gradient learning but can decrease the variance of the updates \citep{sutton2018reinforcement}.

The gradient can be computed as follows:
\begin{equation}
\begin{split}
\frac{\partial}{\partial W^\mathrm{rec}_{ab}} 
\ln \pi(\mathbf{h}^t | \mathbf{h}^{t-1},\mathbf{W}^\mathrm{rec})
&= -\frac{1}{2 \sigma^2_{\mathrm{rec}}}
\frac{\partial}{\partial W^\mathrm{rec}_{ab}}
\big(\mathbf{h} - \bm{\mu}(t,\mathbf{W}^\mathrm{rec})\big)^2 \\
&= \frac{1}{\sigma^2_{\mathrm{rec}}} \sum_i 
\left[ h_i - \mu_i(t, \mathbf{W}^\mathrm{rec}) \right]
\frac{\partial}{\partial W^\mathrm{rec}_{ab}}
\mu_i(t, \mathbf{W}^\mathrm{rec}) \\
&= \frac{1}{\sigma^2_{\mathrm{rec}}} \sum_i \xi_i^t 
\frac{\partial}{\partial W^\mathrm{rec}_{ab}}
\mu_i(t, \mathbf{W}^\mathrm{rec}) \\
&= \frac{1}{\tau\sigma^2_{\mathrm{rec}}} \xi_a^t
\phi'\left( u_a^t\right) h_b^{t-1}.
\end{split}
\end{equation}
In order to address temporally delayed credit assignment, we can additionally incorporate an eligibility trace in the gradient appearing in \eqref{eq:rl_update0}, replacing $\nabla\ln\pi \longrightarrow \overline{\nabla\ln\pi}$, where the bar denotes low-pass filtering \cite{sutton2018reinforcement}. This allows credit for rewards at time $t$ to be assigned to the RNN activity at earlier time steps. This leads to the following update rule:
\begin{equation}\label{eq:rl_update}
    \Delta W^\mathrm{rec}_{ab} 
    = \eta (R^t - \bar{R}^t) q^t_{ab},
\end{equation}
where the eligibility trace is given by
\begin{equation}
    q^t_{ab} 
    = \left( 1 - \frac{1}{\tau_e} \right) q_{ab}^{t-1}
    + \frac{1}{\tau_e} \xi_a^t \phi'(u_a^t) h_b^{t-1},
\end{equation}
with $q_{ab}^0 = 0$.
In our simulations, we set the timescale for the eligibility trace to be equal to the network time constant, i.e.~$\tau_e = \tau$.


In order to compute the expected change $\Delta \mathbf{F}^\mathrm{pred}$ in the RNN flow fields, we next wish to compute the expectation of the weight update \eqref{eq:rl_update} by averaging over noise. 
As in the SL case from the previous section, we assume that the RNN is linear and switch to the continuous-time limit. In this case, the eligibility trace is given by
\[q_{ij}^t = \int_0^t e^{(s-t)/\tau_e} \xi_i^s h_j^s \frac{ds}{\tau_e}.\]

The reward $R^t$ itself is given by
\begin{equation}
R^t = -|\boldsymbol{\epsilon}^t|^2 = -|\mathbf{y}^{*t}|^2 - |\mathbf{y}^t|^2 + 2 \mathbf{y}^{*t} \cdot \mathbf{y}^t.
\end{equation}
We assume that the expected reward $\bar R^t$ is independent of the noise $\xi$, so it doesn't contribute to $\langle \Delta W^\mathrm{rec} \rangle$.
Then we can use Wick's theorem to calculate the expected weight update:
\begin{equation}
\begin{split}
\langle \Delta W^\mathrm{rec}_{ij} \rangle &= \eta \sum_t\sum_k \int_0^t e^{(s-t)/\tau_e} \left\langle (2y^*_k y^t_k - y^t_k y^t_k)\, \xi^s_i\, h^s_j\right\rangle \frac{ds}{\tau_e}\\
&=\eta \sum_t\sum_k \int_0^t e^{(s-t)/\tau_e} \left( 2\langle y_k^* h_j^s \rangle - 2 \langle y_k^t h_j^s \rangle \right) \langle \xi_i^s y_k^t \rangle \frac{ds}{\tau_e}\\
&=2\eta \sum_t \sum_{k,l} W^{\text{bmi}}_{kl}\int_0^t e^{(s-t)/\tau_e} \left( \langle y_k^* h_j^s \rangle - \langle y_k^t h_j^s \rangle \right) \langle \xi_i^s h_l^t \rangle \frac{ds}{\tau_e}.
\end{split}
\end{equation}

For the linear network we are considering, we have $\langle \xi_i^s h_l^t \rangle = \sum\limits_m \Sigma_{im} \left(e^{\mathbf{W}(t-s)}\right)_{lm}$, where $\boldsymbol{\xi}^s \sim \mathcal{N}(0,\boldsymbol{\Sigma})$ and $\mathbf{W}= - \mathbb{I} + \mathbf{W}^\mathrm{rec} + \mathbf{W}^{\text{fb}} \mathbf{W}^{\text{bmi}}$. Then
\begin{equation}\label{eq:dW_avg_RL}
\langle \Delta W^\mathrm{rec}_{ij} \rangle = 2 \eta \sum_t\sum_{k,l,m} W^{\text{bmi}}_{kl}\Sigma_{im} \int_0^t e^{\frac{s-t}{\tau_e}} \left(e^{\mathbf{W}(t-s)}\right)_{lm}\left( \langle \epsilon^t_k \rangle \langle h^s_j \rangle - \sum_r W^{\text{bmi}}_{kr} Q^{t,s}_{rj}\right) \frac{ds}{\tau_e}.
\end{equation}
In the case where $\tau_e$ is unknown (as it may be in experimental data) or sufficiently small, this expression can be simplified by taking the limit $\tau_e \to 0$, leading to
\begin{equation}\label{eq:dW_avg_RL_simple}
\langle \Delta W^\mathrm{rec}_{ij} \rangle 
= 2 \eta \sum_t\sum_{k,l} W^{\text{bmi}}_{kl}\Sigma_{il} \left( \langle \epsilon^t_k \rangle \langle h^t_j \rangle - \sum_r W^{\text{bmi}}_{kr} Q^{t,t}_{rj}\right) .
\end{equation}
As in the SL case, in the limit where the noise is small, the first term in parentheses will be much larger than the second, which can be dropped to obtain a simplified expression.

\section{Simulations}
\label{sec:appendix-simulations}



\subsection{Experimental details}
\label{sec:exp_details}
The code used to run these simulations can be found at \url{www.github.com/jacobfulano/learning-rules-with-bmi}

As stated in the main text, we first pretrain the recurrent weights $\mathbf{W}^{\mathrm{rec}}$ of an RNN with  fixed, random readout weights $\mathbf{W}^{\mathrm{bmi0}}$ to perform a center-out cursor-control task, in which a cursor must be moved to one of four target locations specified by the input to the RNN. We then change the BMI decoder to $\mathbf{W}^{\mathrm{bmi1}}$, make identical copies of the network, and train one with SL and the other with RL. Multiple seeds ($n=4$) were selected for $\mathbf{W}^{\mathrm{bmi1}}$ and $\mathbf{M}$ and applied to network copies in the training phase, and correlation metric results $\mathrm{Corr}(\Delta \mathbf{F}^{\mathrm{obs}}, \Delta \mathbf{F}^{\mathrm{pred}})$ were averaged over these seeds. Error bars are S.E.M. Unless stated otherwise, all simulations involved pretraining.

Simulations for Figures \ref{fig:main-results} used 4 input dimensions, $N=50$ recurrent units, and 2 output dimensions, with a trial duration of 20 timesteps. The scale for the variance of the noise injected at each layer was $\sigma^2_{\mathrm{in}}=0, \sigma^2_{\mathrm{rec}}=0.25,\sigma^2_{\mathrm{bmi}}=0.01$ for input, recurrent, and output units respectively. Constant learning rate for recurrent weights was $\eta^{\mathrm{rec}}=0.1$, and the RNN time constant was set to $\tau=10$. Recurrent weights $\mathbf{W}^{\mathrm{rec}}$ were initialized with $\mathbf{W}^{\mathrm{rec}} \sim \mathcal{N}(0,g^2/\sqrt{N})$ where $g=1.5$. Input weights $\mathbf{W}^{\mathrm{in}}$ and decoder weights $\mathbf{W}^{\mathrm{bmi}}$ were initialized randomly and uniformly over $[-2,2]$ and $[-2/\sqrt{N},2/\sqrt{N}]$ respectively.  For all networks, the activation function was $\phi(\cdot) = \tanh(\cdot)$. Pretraining was run for 2,500 trials using SL, and training was run for 1,500 trials for SL and 15,000 trials for RL. Alignment between $\mathbf{M}$ and $\mathbf{W}^{\mathrm{bmi0}}$ was fixed at 0.5. Input signals for each target consisted of a step function that was 1 for $20\%$ of the trial duration (i.e. 4 timesteps) and 0 for the remainder. Input to the network at each timestep was therefore a 4 dimensional vector with one entry equal to 1 and other entries equal to zero. For both SL and RL algorithms, weights were updated at the end of each trial (i.e. ``offline''). For RL simulations using \eqref{eq:RL_update}, a separate reward baseline $\bar{R}^t$ was kept for each target.


In order to control alignment $\alpha$ between decoder weights $(\mathbf{W}^{\mathrm{bmi}})^\top$ and matrices $\mathbf{M}$, we generated a matrix $\mathbf{M}$ by randomly changing a subset of matrix entries from $(\mathbf{W}^{\mathrm{bmi}})^\top$ such that $\mathrm{sim}(\mathbf{M},(\mathbf{W}^{\mathrm{bmi}})^\top)=\alpha$. Networks trained with SL were able to consistently learn the task with four targets for $\alpha > 0.3$. Throughout this study, analyses were only performed on networks that successfully learned the center-out reach task.

When calculating the flow field metric in \eqref{eq:dF_corr}, ``early'' (before learning) and ``late'' (after learning) blocks consisted of 500 trials. Activity during learning was split into training trials and test trials; predictions of the (direction of) weight change $\Delta \mathbf{W}^{\mathrm{pred}}|_{SL}$ and $\Delta \mathbf{W}^{\mathrm{pred}}|_{RL}$ were constructed using activity $\mathbf{h}^{n,t}$ and error $\boldsymbol{\epsilon}^{n,t}$ from the training trails, and the full metric $\mathrm{Corr}(\Delta \mathbf{F}^{\mathrm{obs}}(\mathbf{h}),\Delta \mathbf{F}^{\mathrm{pred}}(\mathbf{h}))$ was evaluated on activity $\mathbf{h}^{n,t}$ from the test trials.

\begin{figure}[t!]\label{fig:hyperparameters}
  \centering
  \includegraphics[width=0.8\textwidth]{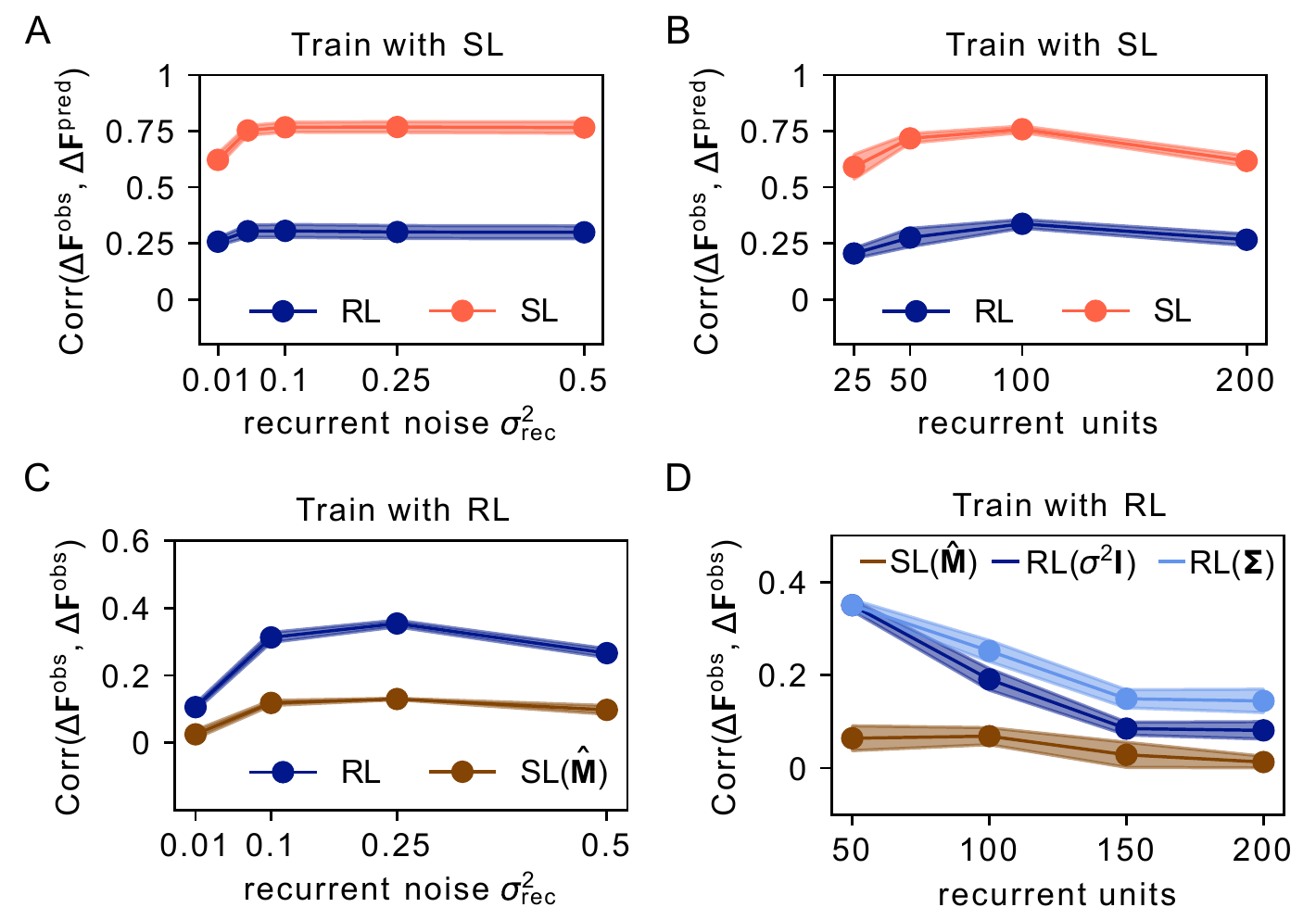}
  \caption{Learning rules are distinguishable across hyperparameter choices. (A) Varying recurrent noise while training with SL for networks with 50 recurrent units. (B) Varying the number of recurrent units while training with SL. Both (A) and (B) were obtained using $\mathrm{sim}(\mathbf{M}, (\mathbf{W}^\mathrm{bmi1})^\top) = 0.5$, with other hyperparameters the same as in Fig.~\ref{fig:main-results}. 
  (C) Varying recurrent noise while training with RL for networks with 50 recurrent units. (D) Varying the number of recurrent units while training with RL. Noise dimension was set to 50, and correlation was calculated for true, low-dimensional noise covariance (light blue), or naive, full dimension estimate of the noise covariance (dark blue). For both (C) and (D), $\mathbf{\hat{M}}$ was randomly sampled such that $\mathrm{sim}(\mathbf{\hat{M}}, (\mathbf{W}^\mathrm{bmi1})^\top) = 0.5$; other hyperparameters are the same as in Fig.~\ref{fig:main-results}.}
\end{figure}

Fig.~\ref{fig:hyperparameters} shows that the flow field correlation metric successfully distinguishes the learning rules across hyperparameters, including recurrent noise $\sigma^2_{\mathrm{rec}}$ and number of recurrent units, for both SL and RL. For networks with a large number of recurrent units, we found that the RL node perturbation algorithm was more effective when the noise was low-dimensional. In Fig.~\ref{fig:hyperparameters}D, therefore, the recurrent noise is 50-dimensional, and isotropic within those dimensions.

For Fig.~\ref{fig:feedback}A, weight mirroring \cite{akrout2019deep} was used as a convenient way to update $\mathbf{M}$ while also learning recurrent weights $\mathbf{W}^{\mathrm{rec}}$ with SL. Network parameters were the same as Fig.~\ref{fig:main-results}, except that $\eta^{\mathrm{rec}}$ was lowered to 0.05 and the weight mirroring learning rate was $\eta^{\mathrm{WM}}=0.001$. The weight mirroring algorithm applied to our context simply correlates presynaptic recurrent noise $\xi_j \sim \mathcal{N}(0, \sigma^2_{\mathrm{rec}})$ with postsynaptic activity $y_i = W^{\mathrm{bmi}}_{ij} \xi_j$ and then updates $\mathbf{M}$ with a update rule $\Delta M_{ji} = \eta^{\mathrm{WM}} \xi_j y_i $. On average, this pushes $\mathbf{M}$ in the direction of $\mathbb{E}[\xi_j y_i ] = \sigma^2_{\mathrm{rec}} W^{\mathrm{bmi}}_{ji}$, i.e. the transpose of the decoder weights $\mathbf{W}^{\mathrm{bmi}}$.

For Fig.~\ref{fig:feedback}C-E, we included driving feedback weights $\mathbf{W}^{\mathrm{fb}} = \gamma \mathbf{M}$ and varied the strength of the driving feedback weights by a scalar factor $\gamma$ between 0.5 and 5. For these simulations, $\mathbf{M}$ was kept fixed at $\mathrm{sim}(\mathbf{M},(\mathbf{W}^{\mathrm{bmi}})^\top)=0.5$. Training ran for 5,000 trials for networks using SL, and 10,000 trials for networks using RL. Network parameters were the same as Fig.~\ref{fig:main-results}, except with $\eta^{\mathrm{rec}}=0.1$ and $\sigma^2_{\mathrm{rec}}=0.1$.



For Fig.~\ref{fig:estimate-m}, the stimulus signal was equal to 1 for the full trial, and the feedback weights were set to $\mathbf{W}^{\mathrm{fb}} = \gamma \mathbf{M}$ with $\gamma=5$. Pretraining lasted for 2,500 trials, while training was set to 1,000 trials and the overlap between $\mathbf{M}$ and $\mathbf{W}^{\mathrm{bmi0}}$ was set to 0.5. Network parameters were the same as Fig.~\ref{fig:main-results}, except with $\eta^{\mathrm{rec}}=1$, and $\sigma^2_{\mathrm{rec}}=0.2$.

For Fig.~\ref{fig:non-iso}, we trained networks to learn a cursor-control task with non-isotropic noise via either SL or RL. In the SL case, we chose a biased credit mapping $\mathbf{M}$ that has partial overlap with the new decoder, with $\mathrm{sim}(\mathbf{M}, (\mathbf{W}^{\mathrm{bmi}1})^\top)=0.6$. Network parameters were the same as Fig.~\ref{fig:main-results}, except with $\eta^{\mathrm{rec}}=0.2$. Training was run for 1,000 trials. The recurrent noise covariance was chosen to be $d$-dimensional, with $\mathrm{rank}(\boldsymbol{\Sigma}) = d$, and isotropic within those dimensions. Simulations were run for $d={5,10,25,50}$ with three seeds for each dimension. The first two of these dimensions were selected to lie in the subspace spanned by $\mathbf{M}$ for the SL-trained RNNs or by $\mathbf{W}^\mathrm{bmi1}$ for the RL-trained RNNs via QR decomposition. Other components of $\boldsymbol{\Sigma}$ were added in random dimensions orthogonal to this subspace and to one another.

In Fig.~\ref{fig-feedforward}, the network size was 20-20-2, and the task was to map $T=5$ random binary input patterns onto random output targets. 
For SL, the parameters were $\eta_\mathrm{SL} = 0.001$, $\sigma = 0.1$,  $N_\mathrm{trials} = 500$, and $N_\mathrm{early} = N_\mathrm{late} = 10$. For RL, the parameters were $\eta_\mathrm{SL} = 0.003$, $\sigma = 0.1$, $N_\mathrm{trials} = 5000$, and $N_\mathrm{early} = N_\mathrm{late} = 100$. In both cases, the results shown in Fig.~\ref{fig-feedforward}C,F were computed by averaging over 100 different networks for each condition.

\subsection{Alternative supervised learning rules: ``Biased'' Backpropagation Through Time (BPTT)}
\label{sec:biased_bptt}

As described in the main text, RFLO \cite{murray2019local} is an approximate gradient-based algorithm with $\Delta \mathbf{W}^{\mathrm{rec}} \approx -\partial L / \partial \mathbf{W}^{\mathrm{rec}}$. The main results of this study depend on the key idea that the matrix $\mathbf{M}$ used in the (supervised) learning rule is not identical to the transpose of the decoder weights $(\mathbf{W}^{\mathrm{bmi}})^\top$. This idea can be applied to BPTT, leading to a learning rule we call ``biased'' BPTT. The standard BPTT update is
\begin{equation}
    \frac{\partial L}{\partial W^{\mathrm{rec}}_{ab}} = - \frac{1}{\tau T} \sum_t z_a^t \phi'(u_a^t) h_b^{t-1}
\end{equation}

\begin{equation}
    z_i^t = \sum_j W^{\mathrm{bmi}}_{ji} \epsilon_j^t +  \left(1-\frac{1}{\tau}\right) z_i^{t+1} + \frac{1}{\tau} \sum_j \phi'(u_j^{t+1}) W^{\mathrm{rec}}_{ji}z_j^{t+1}
\end{equation}

with $u_i^t = \sum_k W^{\mathrm{rec}}_{ik} h_k^{t-1} + \sum_k W^{\mathrm{in}}_{ik} x_k^t  + \sum_k W^{\mathrm{fb}}_{ik} y_k^t $, as in the main text. For a careful comparison of BPTT, RFLO, and other related gradient-based algorithms for training RNNs, see Refs. \cite{murray2019local,marschall2020unified}. 

A \textit{biased} BPTT learning rule according to our framework would simply replace the $(\mathbf{W}^{\mathrm{bmi}})^\top \boldsymbol \epsilon^t$ term with $\mathbf{M} \boldsymbol \epsilon^t$:

\begin{equation}
    z_i^t = \sum_j M_{ij} \epsilon_j^t +  \left(1-\frac{1}{\tau}\right) z_i^{t+1} + \frac{1}{\tau} \sum_j \phi'(u_j^{t+1}) W^{\mathrm{rec}}_{ji}z_j^{t+1}
\end{equation}

\begin{figure}[t!]
  \centering
  \includegraphics[width=\textwidth]{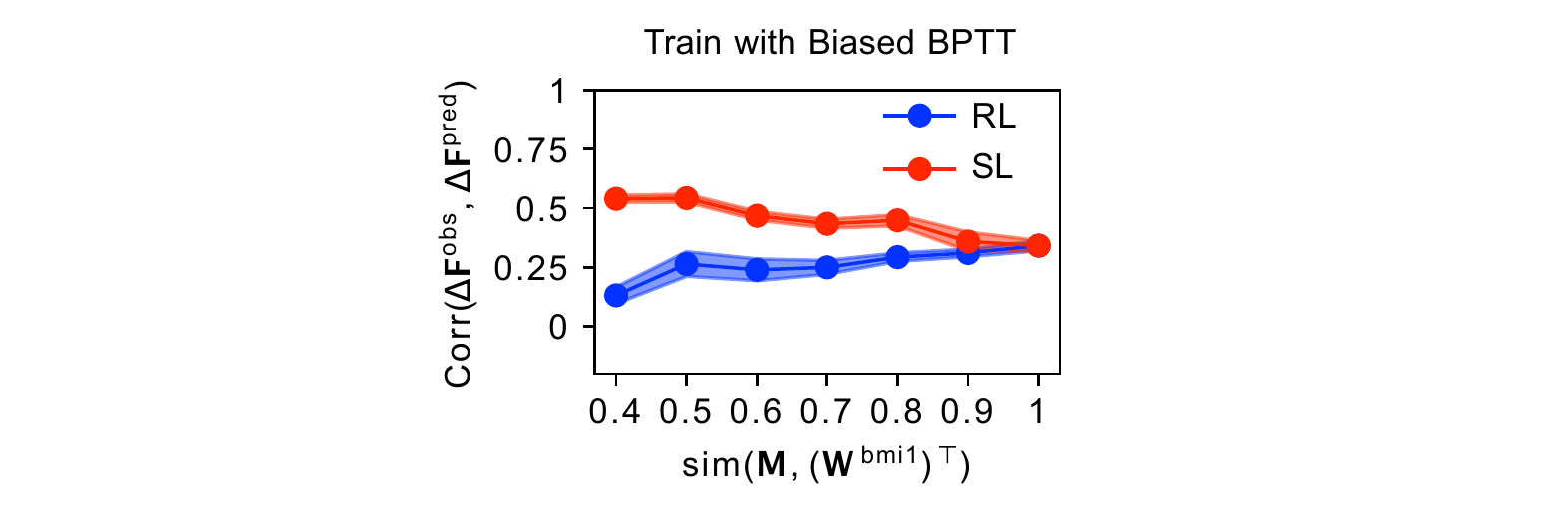}
  \caption{Correlation metric generalizes to other SL algorithms. Networks are trained with a biased form of BPTT, where $(\mathbf{W}^{\mathrm{bmi1}})^\top$ is replaced by  $\mathbf{M}$ in the weight update rule. }
  \label{fig:bptt}
\end{figure}

We show in Fig.~\ref{fig:bptt} that our results hold when using a biased BPTT learning rule instead of the supervised RFLO learning rule. Network parameters and simulation details were the same as Fig.~\ref{fig:main-results}, except that a biased BPTT learning rule was used. For lower similarity between $\mathbf{M}$ and $(\mathbf{W}^{\mathrm{bmi}})^\top$, the correlation metric is able to correctly identify bias in the change in flow field. Compared with the results shown in Fig.~\ref{fig:main-results}, the correlation metric values are somewhat lower for biased BPTT. This is likely because additional nonlocal recurrent terms are contained in the true weight update that are not accounted for in the expression for $\Delta \mathbf{F}^{\mathrm{pred}}_{\mathrm{SL}}$ that uses $\Delta \mathbf{W}^{\mathrm{pred}} |_{\mathrm{SL}}$ from \eqref{eq:W_pred_SL}.


\begin{figure}[h]
  \centering
  \includegraphics[width=\textwidth]{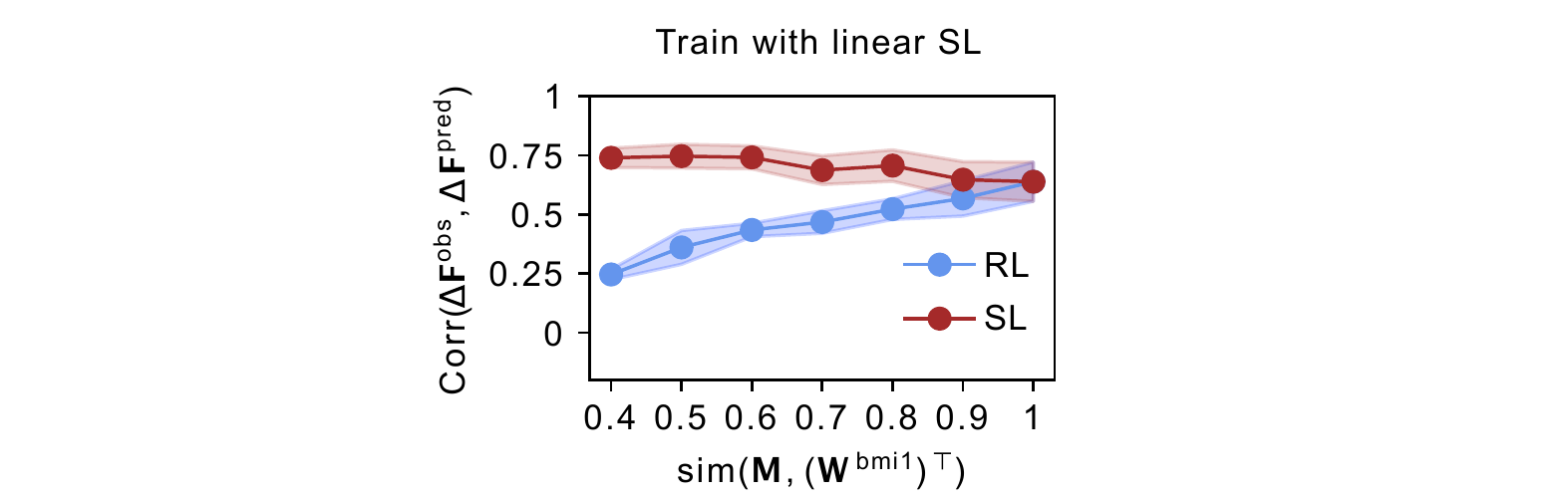}
  \caption{Correlation metric generalizes to linear networks.}
  \label{fig:linear}
\end{figure}

\subsection{Linear RNNs}

We show that the RNN nonlinearity $\phi(\cdot)=\tanh(\cdot)$ is not a crucial architectural choice for our main conclusions. In Fig.~\ref{fig:linear} we apply the same approach (and hyperparameters) as shown in Fig.~\ref{fig:main-results}C to linear RNNs and find similar results.


\begin{figure}[t!]
  \centering
  \includegraphics[width=\textwidth]{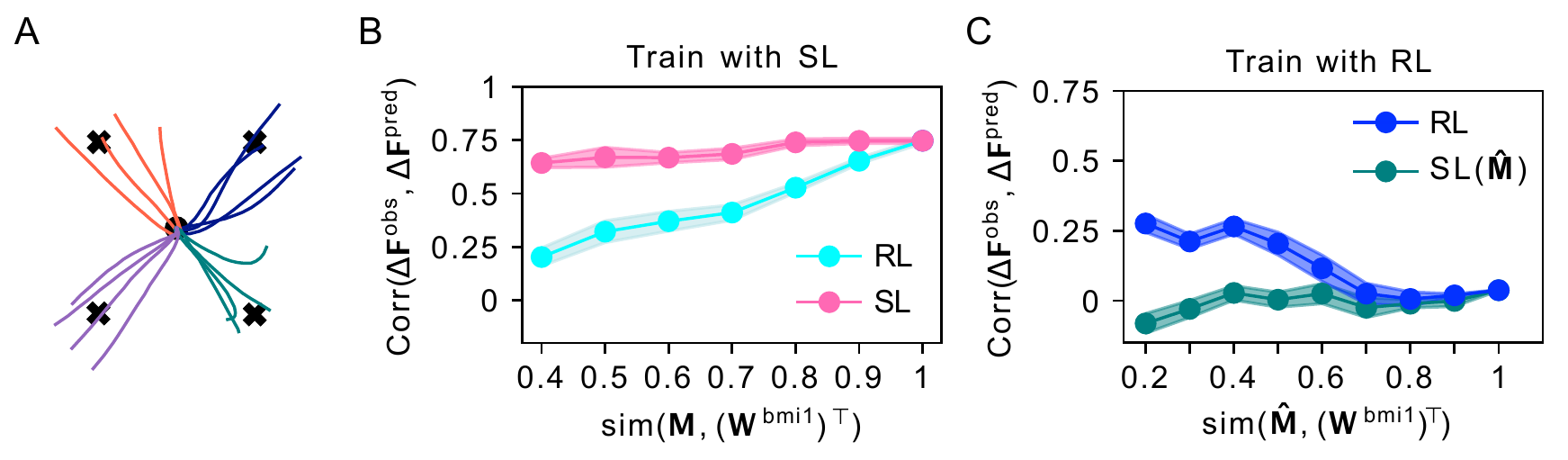}
  \caption{SL and RL are distinguishable with velocity-based cursor control. (A) Example trajectories from a trained RNN in which BMI readout weights map neural activity to cursor velocity.
  (B) $\mathrm{Corr}(\Delta \mathbf{F}^{\mathrm{obs}},\Delta \mathbf{F}^{\mathrm{pred}})$ for RNNs trained with SL to control cursor velocity. (C) $\mathrm{Corr}(\Delta \mathbf{F}^{\mathrm{obs}},\Delta \mathbf{F}^{\mathrm{pred}})$ for RNNs trained with RL to control cursor velocity.}
  \label{fig:velocity}
\end{figure}

\subsection{Velocity-based cursor control}
In the main text, we modeled a task in which a BMI readout maps neural activity directly onto cursor position. While this is a conceptually simple way to illustrate our main ideas, actual BMI experiments more commonly use the readout of neural activity to control cursor \textit{velocity} rather than position (e.g.~Refs.~\cite{sadtler2014neural,athalye2017emergence,oby2019new,sorrell2021brain}). In this section we show that our main results can also be obtained for this case. 

Let the RNN readout $\mathbf{y}^t = \mathbf{W}^\mathrm{bmi} \mathbf{h}^t$ correspond to cursor velocity rather than to cursor position, and let the cursor position be given by $\mathbf{r}^t = (1 - 1/\tau_r) \mathbf{r}^{t-1} + \mathbf{y}^t / \tau_r$. For simplicity, we take $\tau_r = \tau$ in our simulations. Let the target velocity at time $t$ be given by $\mathbf{y}^{*t} = \mathbf{r}^* - \mathbf{r}^t$, where $\mathbf{r}^*$ is the target position, and the error be given by $\boldsymbol{\epsilon}^t = \mathbf{y}^{*t} - \mathbf{y}^t$. 

Using velocity-based cursor control leads to smoother cursor trajectories, as shown in Fig.~\ref{fig:velocity}A. Figures \ref{fig:velocity}B-C show that the learning rules can be correctly identified in RNNs that are trained under these assumptions using either SL or RL.


\begin{figure}[h]
  \centering
  \includegraphics[width=\textwidth]{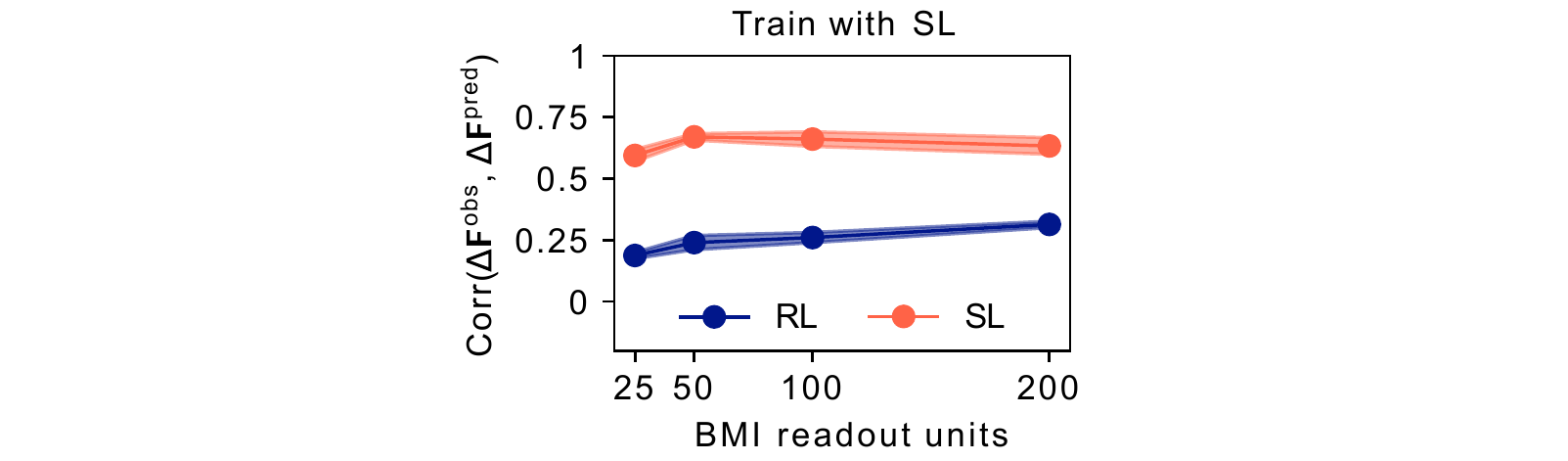}
  \caption{Correlation metric applied to networks with 200 recurrent units. BMI decoders only read out activity from a subset of neurons in the recurrent population.}
  \label{fig:bmi-units}
\end{figure}

\subsection{Learning with a BMI that samples from a subset of the neural population}
\label{sec:update_direction}
The simulations in this study have assumed that the decoder $\mathbf{W}^{\mathrm{bmi}}$ reads out the neural activity from all the neurons in the RNN. 
While this assumption is not realistic with respect to neuroscience experiments, we show here that it does not affect our conclusions.

We ran simulations for RNNs with 200 recurrent units using SL and varied the number of units read out by the decoder between 25 and 200, setting the decoder weights of all non-readout units to zero. 
After pretraining the RNN, the decoder $\mathbf{W}^{\mathrm{bmi1}}$ was randomly selected such that $\mathrm{sim}(\mathbf{W}^{\mathrm{bmi}0}, \mathbf{W}^{\mathrm{bmi}1})=0.5$, while reading out from the same units that were read out from during pretraining. 
The credit assignment mapping $\mathbf{M}$ was randomly chosen such that $\mathrm{sim}(\mathbf{M}, (\mathbf{W}^{\mathrm{bmi}1})^\top)=0.5$, and was not necessarily restricted to the same subset of readout units. 
This was repeated with different numbers of readout units, each across 5 random seeds. 
Network and training hyperparameters were otherwise the same as in Figures \ref{fig:main-results} and \ref{fig:hyperparameters}.
Fig.~\ref{fig:bmi-units} shows that, in this more realistic scenario where a BMI only decodes a subset of the neurons in the neural population, our correlation metric continues to distinguish between the SL and RL training algorithms.

\subsection{Weight updates are distinct for biased SL and RL}

In order to build an intuition for how the SL and RL rules affect weight updates, we analyzed these updates directly (rather than the changes in flow fields, which was the focus in the main text).


In the SL case we use \eqref{eq:W_pred_SL}:
\begin{equation*}
     \Delta W_{ij}^{\mathrm{pred}}|_{\mathrm{SL}} = \sum_t^T \sum_k M_{ik} \epsilon_k^t  h^t_j.
\end{equation*}
In the RL case we use \eqref{eq:W_pred_RL}:
\begin{equation*}
     \Delta W_{ij}^{\mathrm{pred}}|_{\mathrm{RL}} =  \sum_t^T \sum_k W^{\mathrm{bmi}}_{kl} \Sigma_{il} \epsilon_k^t  h^t_j,
\end{equation*}
with isotropic noise $\boldsymbol{\Sigma} = \sigma^2_{\mathrm{rec}} \mathbf{I}$.

\begin{figure}[t!]\label{fig:weight-alignment}
  \centering
  \includegraphics[width=\textwidth]{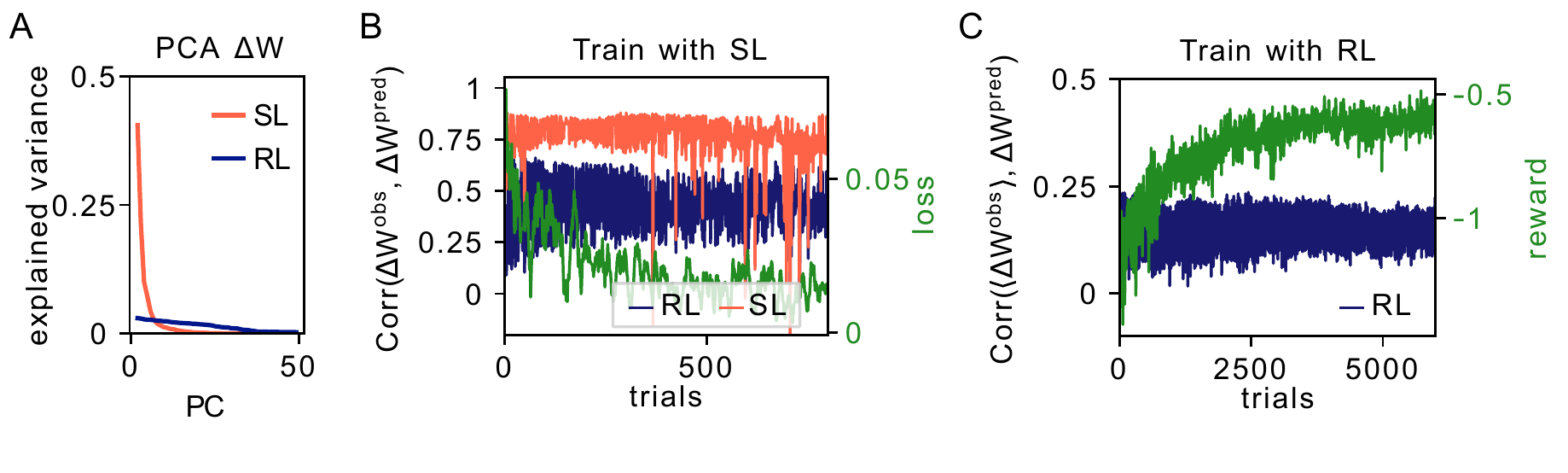}
  \caption{Observed weight updates follow the predicted update direction.  (A) PCA on the weight changes for a network trained with SL  (red) and for a network trained with RL (blue). (B) In a network trained with SL, the observed weight updates from individual trials are correlated with $\Delta \mathbf{W}^{\mathrm{pred}}$ predicted by SL (using matrix $\mathbf{M}$).  (C) For a network trained with RL, the observed weight updates averaged over trials are correlated with $\Delta \mathbf{W}^{\mathrm{pred}}$ predicted by RL (using matrix $\mathbf{W}^{\mathrm{bmi}}$). }
\end{figure}

Fig.~\ref{fig:weight-alignment}B shows that, for an RNN trained with SL, the observed weight updates $\Delta \mathbf{W}^{\mathrm{obs}}$ for each trial are highly correlated with the predicted weight updates $\Delta \mathbf{W}^{\mathrm{pred}}|_{\mathrm{SL}}$. This is particularly pronounced early in learning, as the loss is still decreasing. This analysis is applied to a network trained with SL with $\mathrm{sim}(\mathbf{M},(\mathbf{W}^{\mathrm{bmi}})^\top)=0.6$, and is one way of building intuition for the results in Fig.~\ref{fig:main-results}C and the FFCC metric $\mathrm{Corr}(\Delta \mathbf{F}^{\mathrm{obs}},\Delta \mathbf{F}^{\mathrm{pred}})$.

The RL updates look quite different. Principal component analysis on the weight changes for a network trained with RL shows that weight updates are much more spread out across PCs (Fig.~\ref{fig:weight-alignment}A), which also indicates that individual weight updates don't follow one direction in the loss landscape.  In Fig.~\ref{fig:weight-alignment}C,  however, the \textit{average} of the observed weight updates over trials during learning $\langle \Delta \mathbf{W}^{\mathrm{obs}} \rangle$ is correlated with $\Delta \mathbf{W}^{\mathrm{pred}}|_{\mathrm{RL}}$ predicted after each trial throughout learning.




\subsection{Statistical significance of simulations}
\label{sec:covariance_overlap}
\begin{figure}[t!]\label{fig:t-test}
  \centering
  \includegraphics[width=0.9\textwidth]{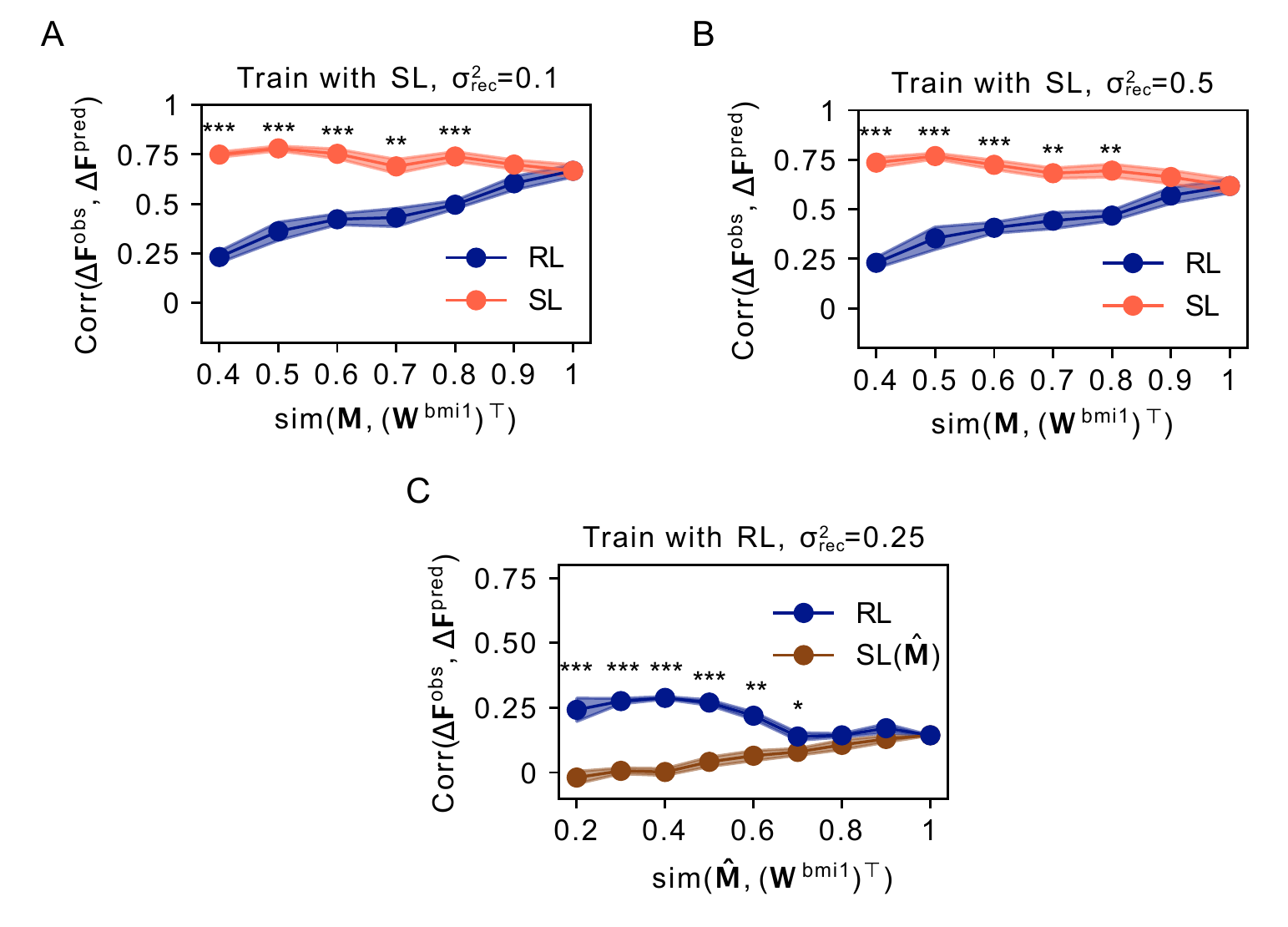}
  \caption{Statistical significance of the correlation metric. (A) The std of the recurrent noise was set to 0.1, with the simulations being otherwise identical to the simulations in Fig. \ref{fig:main-results}C. (B) Same as in (A), but with the recurrent noise set to 0.5. (C) RL simulation data from Fig. \ref{fig:main-results}F. Two sample t-test: (*) indicates $p<0.05$, (**) indicates $p<0.01$, and (***) indicates $p<0.001$.}
\end{figure}

To confirm the statistical significance of our results and their dependence on the level of noise in the RNN, we repeat the simulations from Figure \ref{fig:main-results} with different levels of noise in Figure \ref{fig:t-test} and show that predictions of the two learning rules are statistically distinct when the alignment between the BMI decoder and the credit assignment weights is sufficiently small.




\subsection{Change of neural activity manifold with training}

\begin{figure}[t!]\label{fig:sup-cov-matrix}
  \centering
  \includegraphics[width=1\textwidth]{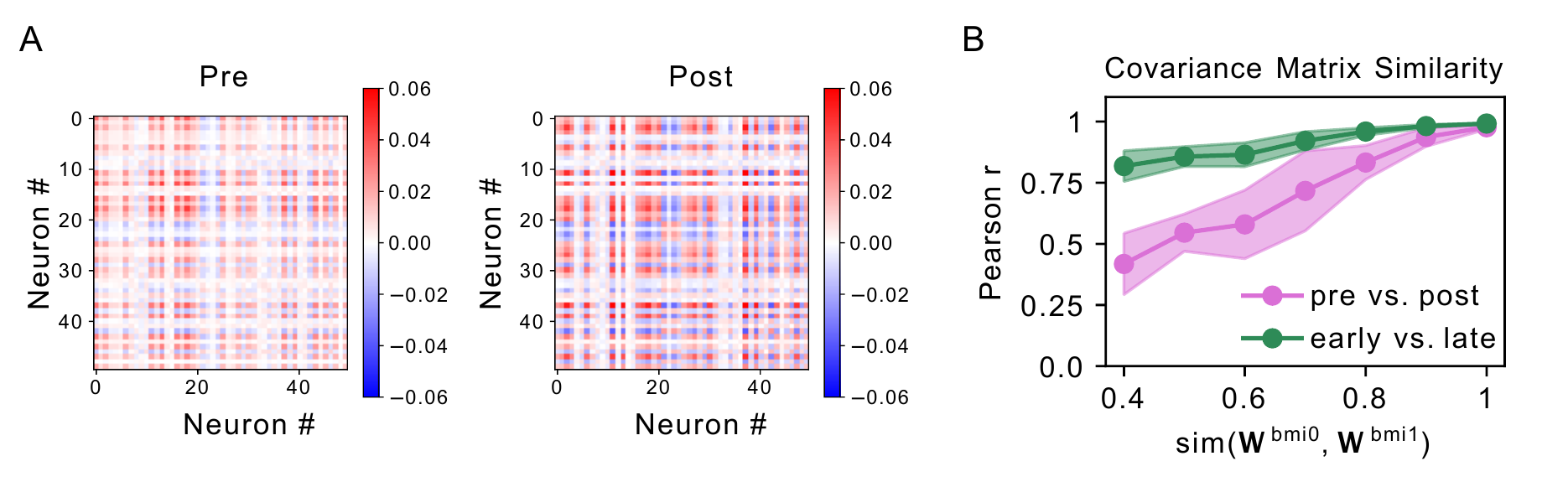}
  \caption{Change of neural activity manifold with training. (A) Example covariance matrices for activity before (``pre'') and after (``post'') learning a new decoder via SL. For these examples, $\mathrm{sim}(\mathbf{W}^{\mathrm{bmi}0}, \mathbf{W}^{\mathrm{bmi}1})=0.8$ (B) Pearson $r$ for pairs of covariance matrices as a function of the alignment between $\mathbf{W}^{\mathrm{bmi}0}$ and  $\mathbf{W}^{\mathrm{bmi}1}$. Magenta line shows the correlation of covariance matrices for activity before (``pre'') and after (``post'') learning a new decoder via SL. Green line shows the Pearson correlation of covariance matrix pairs for activity in the first half (``early'') and second half (``late'') of learning a new decoder via SL (n=3 seeds; error bars show standard deviation).}
\end{figure}

In order to predict the flow field change for a point $\mathbf{h}$ in neural activity space, the direction of weight change is predicted via \eqref{eq:W_pred_SL} and \eqref{eq:W_pred_RL} by sampling activity from trials during training.
In order to verify that there is significant overlap of the distributions from which $\mathbf{h}$ is sampled at different points during training and retraining, we compared the covariance matrices of the RNN activity (Figure \ref{fig:sup-cov-matrix}A) from early trials vs.~later trials with the same decoder (Figure \ref{fig:sup-cov-matrix}B, green curve) or trials after retraining on a new decoder (Figure \ref{fig:sup-cov-matrix}B, magenta curve).
These results show that, although the alignment between the neural activity manifolds before vs.~after retraining with a new decoder decreases as the old and new decoders become less similar, a significant degree of overlap remains even when the decoders are highly dissimilar.

\end{document}